%% file: arxiv.tex
\documentclass{article}

\usepackage[final]{neurips_2025}

\usepackage[utf8]{inputenc} 
\usepackage[T1]{fontenc}    
\usepackage{hyperref}       
\usepackage{url}            
\usepackage{booktabs}       
\usepackage{amsfonts}       
\usepackage{nicefrac}       
\usepackage{microtype}      
\usepackage{xcolor}         

\usepackage{amsmath}
\usepackage{graphicx}
\usepackage{tikz}
\usepackage{xcolor}
\usepackage{fontawesome5}
\usepackage{pifont}
\usepackage{dsfont}
\usetikzlibrary{shapes.geometric, arrows.meta, positioning}
\usepackage{multirow}
\usepackage{wrapfig}
\usepackage{booktabs}
\usepackage{caption}
\usepackage{graphicx}
\usepackage{subcaption}

\input{macros}

\title{Policy Optimized Text-to-Image Pipeline Design}

\author{Uri Gadot\textsuperscript{1,2}
\and
Rinon Gal \textsuperscript{2}
\and
Yftah Ziser \textsuperscript{2,3}
\and
Gal Chechik \textsuperscript{2}
\and
Shie Mannor \textsuperscript{1,2}\and
\textsuperscript{1}Technion \quad \textsuperscript{2}NVIDIA Research
\quad \textsuperscript{3} University of Groningen
}

\begin{document}

\maketitle

\input{sections/0_abstract}
\input{sections/1_intro}

\input{sections/2_prelimenaries}

\input{sections/jpg_sections/method}
\input{sections/jpg_sections/experiments}
\input{sections/5_discussion}

\bibliography{neurips_2025}
\bibliographystyle{plain}

\appendix
\include{sections/jpg_sections/appendix}

\end{document}

%% file: macros.tex
\usepackage{amsmath}
\usepackage{comment}
\usepackage{multirow,bigdelim}
\usepackage{lipsum}
\usepackage{array}
\usepackage{wrapfig}

\usepackage{adjustbox}
\usepackage[percent]{overpic}
\usepackage{makecell}
\usepackage{xcolor}
\usepackage{soul}
\usepackage{amsfonts}
\usepackage{dsfont}
\usepackage{multirow}

\usepackage[T1]{fontenc}   
\usepackage{lmodern}       
\usepackage{microtype}     

\usepackage{xurl}     
\usepackage{hyperref} 
\makeatletter
\newcommand{\settitle}{\@maketitle}
\makeatother

\newcolumntype{C}[1]{>{\centering\let\newline\\\arraybackslash\hspace{0pt}}m{#1}}

\newif\ifdraft
\drafttrue

\ifdraft
\definecolor{darkpink}{rgb}{0.561, 0.282, 0.427}

\newcommand{\ugc}[1]{{\color{red}[\textbf{UG:} #1]}}
\newcommand{\rgc}[1]{{\color{purple}[\textbf{RG:} #1]}}
\newcommand{\yzc}[1]{{\color{blue}[\textbf{YZ:} #1]}}
\newcommand{\smc}[1]{{\color{teal}[\textbf{SM:} #1]}}

\newcommand\galGC[1]{\textcolor{reddish}{[GC: #1]}}

\newcommand{\drop}[1]{}

\else
\newcommand{\ugc}[1]{}
\newcommand{\rgc}[1]{}
\newcommand{\yzc}[1]{}
\newcommand{\gcc}[1]{}
\newcommand{\smc}[1]{}
\newcommand{\galGC}[1]{}

\fi

\newcommand{\ourmethod}{FlowRL}

\makeatother

\usepackage[bottom]{footmisc}
\usepackage{arydshln}

\makeatletter
\def\blfootnote{\xdef\@thefnmark{}\@footnotetext}
\makeatother



%% file: sections/0_abstract.tex
\begin{abstract}


    Text-to-image generation has evolved beyond single monolithic models to complex multi-component pipelines. These combine fine-tuned generators, adapters, upscaling blocks and even editing steps, leading to significant improvements in image quality. However, their effective design requires substantial expertise. Recent approaches have shown promise in automating this process through large language models (LLMs), but they suffer from two critical limitations: extensive computational requirements from generating images with hundreds of predefined pipelines, and poor generalization beyond memorized training examples.
    We introduce a novel reinforcement learning-based framework that addresses these inefficiencies. Our approach first trains an ensemble of reward models capable of predicting image quality scores directly from prompt-workflow combinations, eliminating the need for costly image generation during training. We then implement a two-phase training strategy: initial workflow vocabulary training followed by GRPO-based optimization that guides the model toward higher-performing regions of the workflow space. Additionally, we incorporate a classifier-free guidance based enhancement technique that extrapolates along the path between the initial and GRPO-tuned models, further improving output quality.
    We validate our approach through a set of comparisons, showing that it can successfully create new flows with greater diversity and lead to superior image quality compared to existing baselines. 
\end{abstract}

%% file: sections/1_intro.tex
\section{Introduction}
Recent advancements in generative AI have significantly improved the quality and diversity of text-to-image generation. Early models relied on monolithic architectures, where a single neural network directly translated textual prompts into visual outputs. However, as the field matured, it became clear that combining multiple specialized components—such as fine-tuned diffusion models, super-resolution modules, or specialized embeddings, into more sophisticated workflows leads to superior image quality and greater creative control \cite{cao2024ai,perez2024ai,yao2024improving}. This shift from monolithic models to modular workflows has been supported by user-friendly platforms such as ComfyUI\footnote{\url{https://www.comfy.org/}}, a popular open-source tool that allows users to visually construct complex generative pipelines through interconnected nodes represented in JSON format. ComfyUI has rapidly gained popularity due to its intuitive node-based interface, enabling users to assemble diverse generative models (e.g., Stable Diffusion, ControlNet, LoRAs) into flexible workflows tailored to specific image-generation tasks. Despite its accessibility, designing effective workflows remains challenging due to the vast space of possible component combinations and their prompt-dependent effectiveness. Consequently, crafting high-quality workflows typically requires considerable expertise and manual experimentation.

To address this challenge, recent work introduced ComfyGen \cite{gal2024comfygen}, which uses large language models (LLMs) to automate the construction of prompt-adaptive workflows within ComfyUI. 
However, a key limitation of ComfyGen was its inability to generate genuinely novel workflow structures. At its core, their approach required synthesizing images using an extensive collection of pre-defined workflows and prompts, an expensive process limiting their training set's size. Constrained by this small set, their approach essentially learned a classifier over existing flows rather than synthesizing original graph topologies or selecting novel model combinations. This limitation significantly constrains the potential creativity and adaptability of automated workflow generation systems and, as we later show --- may also limit their downstream performance.


In parallel, reinforcement learning (RL) has emerged as a powerful paradigm for fine-tuning large language models (LLMs), enabling them to optimize their outputs directly based on reward signals derived from human preferences or other evaluative metrics. Techniques such as Reinforcement Learning from Human Feedback (RLHF) have demonstrated remarkable success in aligning model behaviors with human expectations by iteratively refining model parameters based on explicit reward feedback. Furthermore, recent developments like Group Relative Policy Optimization (GRPO) introduced memory-efficient RL algorithms capable of optimizing policies without separate value functions, making them particularly suitable for complex sequential decision-making tasks. Building on these advancements, we propose \ourmethod{}, a novel extension that integrates reinforcement learning into the workflow prediction framework to overcome its originality limitations. Specifically, we formulate workflow generation as an RL problem where an LLM-based policy sequentially constructs workflow graphs by selecting nodes and connections conditioned on textual prompts. To efficiently guide this process without incurring prohibitive computational costs associated with direct image generation for each candidate workflow during training, we introduce a surrogate reward model trained to predict image quality scores directly from prompts and workflow structures.


Finally, we adopt GRPO combined with per-token reward attribution mechanisms to provide granular feedback during policy updates. This affords our RL agent greater precision in identifying decisions within a generated workflow that contribute positively or negatively toward overall image quality.

In summary, our contributions are as follows 
\begin{itemize}
    \item We introduce ComfyGen-RL, the first RL-based approach for generating genuinely novel ComfyUI workflows tailored to align with human preference feedback.
    \item We propose a surrogate human-preference reward model enabling efficient RL training without computationally expensive image generations.
    \item We integrate GRPO with per-token reward attribution for stable and memory-efficient policy optimization.

\end{itemize}


Through these innovations, \ourmethod{} significantly advances automated workflow generation capabilities, enabling richer creativity and greater adaptability in text-to-image synthesis pipelines.



%% file: sections/2_prelimenaries.tex
\section{Related Work}

\textbf{Workflow Generation}



A recent line of research explores the use of compound systems, where multiple models or modules are chained together, often yielding superior performance compared to isolated models. These multi-component systems have been applied across fields ranging from programming challenges~\citep{alpha2024alphacode2} and olympiad-level mathematics~\citep{trinh2024solving} to medical diagnostics~\citep{nori2023can} and video generation~\citep{yuan2024mora}. However, building compound systems presents significant challenges. Models must be chosen not only for their individual strengths, but also for their ability to complement each other. Moreover, the parameters of the different components should be selected with the entire system in mind. To address these difficulties, recent work has explored meta-optimization frameworks, where the structure and parameters of entire pipelines are automatically tuned for downstream performance~\citep{khattab2023dspy}. Others have adopted graph-based architectures allowing dynamic reconfiguration of component interactions~\citep{zhuge2024gptswarm}. 

In the realm of text-to-image generation, recent work explores the use of pipelines using agentic systems~\cite{wang2024genartist,xue2024comfybenchbenchmarkingllmbasedagents,huang2025comfygptselfoptimizingmultiagentcomprehensive}, genetic algorithms~\cite{sobania2024comfygiautomaticimprovementimage} or by fine-tuning LLMs using large flow datasets tagged with human preference scores~\cite{gal2024comfygen}. Although the human preference-based framework has shown promising results, it relies on creating and ranking images using large sets of flows. This, in turn, leads to challenges in effectively scaling the dataset and to a lack of ability to synthesize unseen flows at inference time. Our work aims to address this challenge by leveraging a policy-optimization approach for more effective exploration of the flow parameter space, coupled with a surrogate reward function which avoids the need to generate and rank a large set of images.


\textbf{Fine-Tuning LLMs with RL:}
Reinforcement learning (RL) has become increasingly central to the development of large language models (LLMs), playing a key role in aligning model outputs with user preferences and enhancing task-specific capabilities. A prominent example is Reinforcement Learning from Human Feedback (RLHF) \citep{ouyang2022training}, which fine-tunes models using reward signals derived from human preferences to better align with communicative goals and social norms \citep{dai2023safe, ji2023ai}. Beyond alignment, RL has shown promise in improving LLMs' performance on domains requiring precise reasoning, such as mathematics \citep{uesato2022solving, wang2023math, luo2023wizardmath} and code generation \citep{le2022coderl, li2024falcon}. Recently, \cite{shao2024deepseekmath} proposed Group Relative Policy Optimization (GRPO) as a scalable alternative to Proximal Policy Optimization (PPO). GRPO removes the need for a critic model by optimizing contrastive objective based on intra-group ranking, yielding better sample efficiency, improved stability, and reduced computational complexity \citep{mroueh2025reinforcement, sane2025hybrid}. GRPO-trained LLMs demonstrated state-of-the-art performance in mathematical problem solving and code generation, highlighting its effectiveness on tasks requiring structured reasoning and adherence to correctness \citep{shao2024deepseekmath}. 

\paragraph{Improving Text-to-Image Generation Quality}
The rapid adoption of text-to-image models~\citep{rombach2021highresolutionLDM,nichol2021glide,ramesh2022hierarchical,esser2024scaling,podell2024sdxl} has led to many research efforts focused on improving their image quality and better matching human preferences. Some works focus on inference-time modifications, either optimizing noise seeds towards better behaving regions of the diffusion space~\cite{eyring2024reno,qi2024not} or applying self-guidance and frequency-based modulations~\cite{hong2023improving,si2024freeu,luo2024freeenhance} to the generated features.

More commonly, models are tuned to provide better quality outputs. This is often done through carefully selected high-quality datasets or better captioning methods~\citep{dai2023emu,betker2023improving,segalis2023picture}. Another approach uses reward models~\citep{kirstain2023pickapic,wu2023human,xu2024imagereward,lee2023aligning} to guide the generation process. These reward models can be used with reinforcement learning~\citep{black2024training,deng2024prdp,fan2024reinforcement,zhang2024large}, or through direct optimization~\citep{clark2024directly,prabhudesai2023aligning,wallace2024diffusion}. 

Finally, recent methods explore the use of LLMs to improve text-to-image generation~\cite{yang2024mastering}, commonly by using them to construct workflows featuring multiple models or chained editing tools~\cite{wang2024genartist,sobania2024comfygiautomaticimprovementimage,gal2024comfygen}. Our work similarly uses LLMs to construct workflows, but better aligns them to human preferences through the use of reward models coupled with a reinforcement-learning feedback mechanism.

%% file: sections/jpg_sections/method.tex
\begin{figure}[ht]
\includegraphics[width=1.\linewidth]{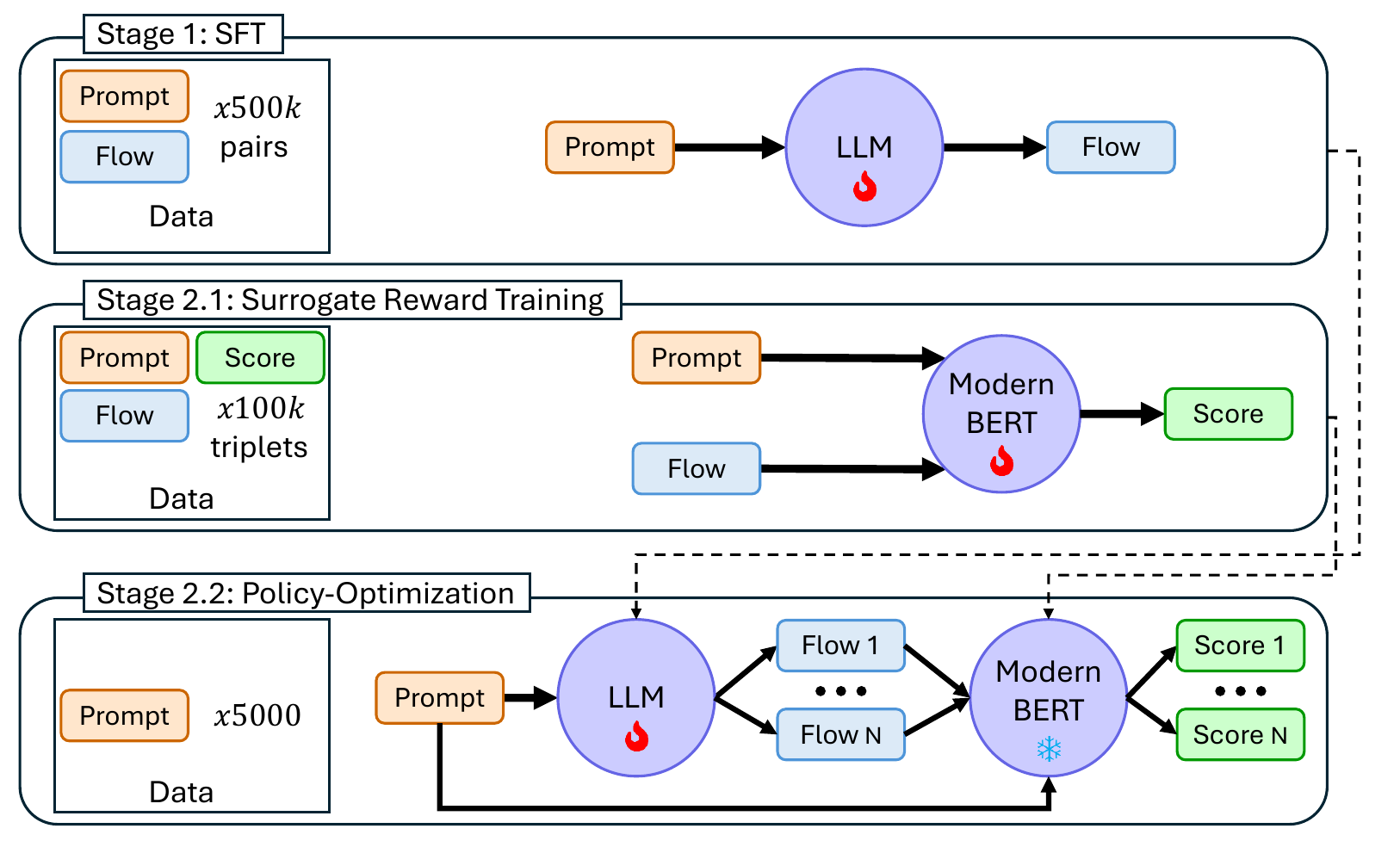}
\caption{Pipeline overview. Step 1: Finetune LLM for general flow generation (SFT, 500K prompt-flow pairs). Step 2.1: Train reward model (100K prompt-flow-score triplets). Step 2.2: Optimize for quality using GRPO. \textcolor{red}{\faFire} = learning, \textcolor{cyan}{\ding{100}} = frozen.}
\label{fig:method}
\end{figure}
\section{Methodology}

Our goal is to enable efficient training of a human-preference based, prompt-to-workflow prediction system. Ideally, this system should be able to innovate and produce novel, unseen flows. Prior work struggled with this aspect, primarily due to their reliance on scoring images generated with a large set of fixed flows, whose parameters were sampled uniformly from a predefined set of options. To overcome this hurdle, we propose a two-phase training strategy. In the first, we pre-train on a large set of un-scored flows. This avoids the need to generate and score images, allowing us to use a much larger set to teach the LLM the structure of flows and the available components. Then, we perform a second tuning stage, where we leverage human-preference predictor models jointly with recent reinforcement-learning ideas (GRPO~\cite{shao2024deepseekmath}) to drive the model towards better-performing subsets of the flow space. As training progresses, more samples are drawn from these regions, and hence, less computation is wasted on inefficient exploration. 

However, generating and scoring images during LLM training is itself a costly process, which requires an order of a minute for every training step. Hence, we draw on ideas from the autonomous driving literature, where costly simulations are often replaced by faster predictors trained to replicate simulation outputs \cite{beglerovic2017testing,joshi2022gas,kalweit2022deep}. Here, we apply this idea by learning surrogate reward models that predict the final image score directly from the prompt and workflow pair. Notably, prior work has observed that such surrogates are susceptible to reward-hacking solutions \cite{gao2023scaling,singhal2023long,wen2024language}. Motivated by findings that ensembles can mitigate reward hacking \cite{coste2023reward,zhai2023uncertainty}, we train an ensemble of such models and use their variance as a measure of uncertainty, allowing us to filter out samples that optimize for any individual surrogate reward model. Below we present these core components in greater detail and provide an overview of additional design choices or components that allow us to increase efficiency further or refine our results. An overview of our training pipeline is shown in Figure \ref{fig:method}.

\subsection{Training Data}
To train our model we use the flow and prompt dataset of ComfyGen~\cite{gal2024comfygen}. This set contains 33 human-created flows that define an overall graph structure, further augmented by randomly sampling novel parameter choices for existing blocks such as different base models, differnet LoRAs, diffusion samplers or even the number of steps and guidance scale.
Since we do not need to score images for our first stage, we can apply more extensive augmentations and create $2,000$ variants from each baseline flow structure (compared with ComfyGen's $100$). The set also contains $10000$ prompts taken from the generation sharing website CivitAI.com. We keep the $500$ prompts used to test ComfyGen as a holdout, and train using the rest.

\subsection{Stage 1: Supervised Fine-Tuning on Flow Dataset}
The first stage involves supervised fine-tuning (SFT) an LLM on a dataset of prompt-flow pairs without explicit score labels. At this stage, our goal is to teach the LLM the appropriate vocabulary and flow structure while maintaining output diversity. Our flow dataset $D_{SFT}$ consists of pairs $(p_i,f_i)$ where $p_i$ represents a randomly sampled prompt and $f_i$ represents a randomly sampled flow. We tune the model to take the sampled prompt $p_i$ and return its matching flow $f_i$. The full LLM query is shown in the supplementary.
After fine-tuning, we evaluate the model’s perplexity on  $D_{SFT}$, achieving a score of $1.9$, which reflects strong alignment with the encoded workflows structural patterns.


\paragraph{Efficient Flow Representation Scheme}

\begin{figure}[ht]
    \centering
    \setlength{\belowcaptionskip}{-6pt}
    \includegraphics[width=0.95\linewidth]{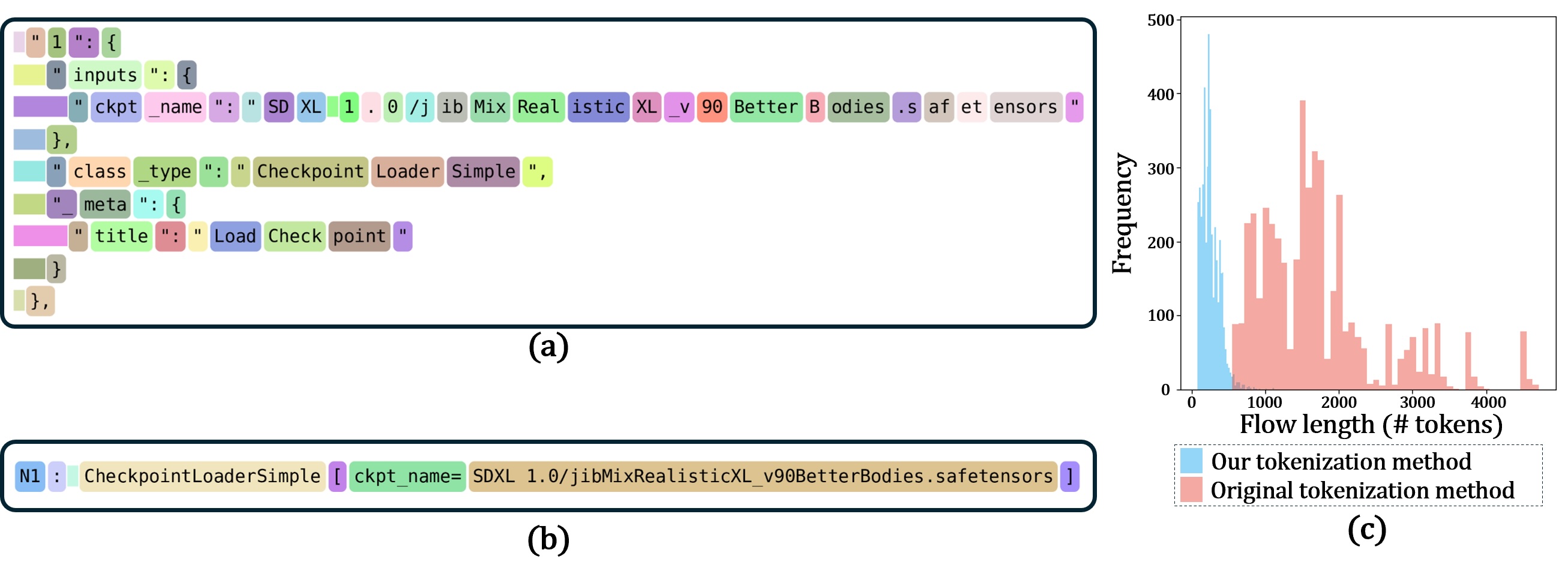}
    \caption{An example of a single ComfyUI node tokenized. \textbf{(a)} displays the original JSON input as tokenized by the standard Llama tokenizer. \textbf{(b)} shows our custom encoding, with introducing additional tokens to explicitly represent relevant components within workflow. Colored segment corresponds to a different token. \textbf{(c)} histogram of flows length (in token) of all training-set}
    \label{fig:encoding}
\end{figure}
While prior work~\cite{gal2024comfygen} directly predicts ComfyUI JSON representations, we note that these JSONs typically contain thousands of tokens, leading to long generation times and increasing memory requirements. An inspection of the tokenized JSONs shows that many tokens are wasted on maintaining the JSON format (e.g., on brackets or quotation marks) or on breaking down model or component names. Hence, to improve training efficiency and reduce token usage, we propose to modify the encoding scheme, using a novel structured representation that captures essential components while reducing token count. Additionally, we introduce specialized tokens to represent key elements of the flow. (e.g. tokens for ComfyUI node names or for model choices). An example of the difference between the two tokenization methods is outlined in Figure~\ref{fig:encoding}.

This new encoding scheme yields significant practical advantages resulting in substantial improvements in both computational efficiency and memory utilization. Quantitatively, the 86.7\% reduction in the average token length ($1500 \rightarrow 200$ tokens per workflow) enabled a $16\times$ batch size increase ($2 \rightarrow 32$ samples/batch) during the first-stage training. Ultimately reaching a $3\times$ time improvement over the original tokenization. These enhancements make it feasible to train complex models and apply memory-intensive algorithms, such as GRPO.


\subsection{Stage 2: Reward-based policy-optimization}

In the second stage, our goal is to tune the workflow-prediction LLM to better align it with flows that produce high quality outputs for a given prompt. To do so, we propose to leverage the recently introduced Group Relative Policy Optimization (GRPO) approach, which estimates advantages by comparing responses within groups of similar prompts, rather than relying on a separate value function. Using GRPO has two main benefits: (1) it eliminates the need to learn a separate value function, enabling better memory utilization during training and (2) its group-based reward normalization encourages greater exploration and diversity in generated workflows.
However, the use of this approach requires us to score and rank the different candidate flows generated for each input prompt at training time. Naively, we could simply generate images with each such flow and score them using the human-preference predictors used by ComfyGen~\cite{gal2024comfygen}. However, for complex flows, creating the images might take an order of a minute, greatly limiting the speed of training. Hence, we propose to avoid this lengthy generation step and instead train a surrogate reward model that will directly estimate the final reward from a pair of prompt and flow inputs.

\paragraph{Surrogate Reward Model Training}
We implement the surrogate reward model $R_{\phi}$ on top of a ModernBert~\cite{warner2024smarter} backbone, with a novel output head trained to map the CLS token into a score. To tune the model, we feed it with strings containing a prompt and flow pair, and task it to predict the human-preference score for the image produced by this pair. For data, we use the ComfyGen dataset $D_{R}$, which contains triplets of prompt $p_i$, flow $f_i$ and score $s_i$. The surrogate's loss is then:
\begin{equation}
    L_{R}(\phi) = \sum_{(p_i,f_i,s_i)\in D_{R}}{MSE(R_{\phi}(p_i,f_i), s_i)} .
\end{equation}

Although the construction of the original ComfyGen dataset still required generating images and scoring them, we find that the surrogate reward is much more sample efficient, performing well with just the $330$ post-augmentation flows of ComfyGen (compared with our own $80k$ unscored flows).


\subsubsection{Component-Aware Hybrid Reward Formulation}
Since downstream flow performance can be heavily influenced by relatively few tokens (model choices, existence of specific blocks), we propose to further refine our surrogate model with a prefix-prediction score that is better able to assign credit to specific components. Specifically, we tune an additional  reward model $R_{\phi}^{prefix}$ to predict the generated image score even when presented only with randomly sampled prefixes of the flow:
\begin{equation}
    L_{R^{pre}}(\phi) = \sum_{(p_i,f_i[1:j],s_i)\in D_{R}}{MSE(R^{pre}_{\phi}(p_i,f_i[1:j]),s_i)} .
\end{equation}

Our final reward design combines these two complementary signals to assign a different reward to each token $t$, depending on both the expected performance of the full flow, as well as a prefix ending with its component:
\begin{equation}
    R(t) = R_\phi(p, f) + \sum_{j=1}^{J}\mathds{1}_{t\in T_j} \cdot R_{\phi}^{pre}(p,f_{1:j}) ,
\end{equation}
where $T$ are the tokens comprising the same flow component as $t$, and we sum over the contribution of the entire component.

\subsubsection{Uncertainty-Aware Reinforcement Learning}
Finally, prior work~\cite{gao2023scaling,singhal2023long,wen2024language} observed that the use of surrogate reward models can lead to reward hacking. To avoid this pitfall, we train an ensemble of $N$ surrogate models $\{R_{\phi_1},R_{\phi_2},...,R_{\phi_N}\}$, each using a different split of our training data. The ensemble provides us with both a more robust mean prediction, as well as with an uncertainty estimate:
\begin{equation}
    \mu(p,f)=\frac{1}{N}\sum_{i=1}^N{R_{\phi_i}(p,f)};\quad
    \sigma(p,f)=\sqrt{\frac{1}{N}\sum_{i=1}^N{(R_{\phi_i}(p,f)-\mu(p,f))^2}} .
\end{equation}

We can then define an uncertainty-aware reward function:
\begin{align*}
R(p,f) = 
\left\{
    \begin {aligned}
         & \mu(p,f) \quad & \sigma(p,f)\leq\tau \\
         & 0\quad & \sigma(p,f)>0
    \end{aligned}
\right.
\end{align*}
where $\tau$ is a threshold parameter. This pessimistic approach assigns zero reward to prompt-flow pairs with high uncertainty, preventing the model from optimizing specific subsets of the reward ensemble, or from drifting to regions where the surrogate's predictions are unreliable.

\subsection{Dual model guidance}
As an additional step, we propose that results may be further improved through the use of a novel inference mechanism inspired by classifier-free guidance (CFG, ~\cite{ho2021classifier}). Specifically, we draw on recent work on image generation~\cite{karras2024guiding} which demonstrate that diffusion models can be guided by extrapolating the predicted scores along the direction from an under-trained version of the model, and the fully trained one. We propose to apply a similar idea here, where we consider both our policy-optimized model ($\mathcal{M}_{GRPO}$, stage 2) and its ``undertrained" SFT version ($\mathcal{M}_{SFT}$, stage 1). At inference time, generations are sampled by interpolating the logits of of both models:
\begin{equation}
    \log{p_{CFG}}(f_j |f_{<j} ,p)= \log{p_{SFT}}(f_j |f_{<j} ,p) + \gamma\big( \log{p_{GRPO}}(f_j |f_{<j} ,p)- \log{p_{SFT}}(f_j |f_{<j} ,p)\big)
\end{equation}
where $p_{CFG}$ represent the sampling distribution, $p_{SFT}$ is the next-token distribution of stage 1 model and $p_{GRPO}$ is the next-token distribution of stage 2 model. Finally, $\gamma\geq0$ controls the guidance strength. Unless otherwise noted, we use $\gamma = 1.5$.

%% file: sections/jpg_sections/experiments.tex
\section{Experiments}


\subsection{Comparisons}

We follow \cite{gal2024comfygen} and compare our approach to a set of baselines across two main metrics: (1) The GenEval~\cite{ghosh2024geneval} benchmark which measures prompt-adherence by using object detection and classification modules to evaluate correct object generation, placement, and attribute binding. (2) Human preference, using the CivitAI prompt-set of ComfyGen~\cite{gal2024comfygen}. For the latter, we evaluate our approach using both an automated preference metric (HPS v2, \cite{wu2023human}) as well as a user study. 

We compare our approach against the following types of baselines: (1) Fixed, monolithic models including: SDXL, popular fine-tuned versions thereof, and SDXL-DPO, which was directly fine-tuned with human preference data. (2) Fixed, popular workflows, where we use the same workflow to generate all images regardless of the prompt. (3) Prior pipeline construction approaches, including agentic workflows that select and use off-the-shelf editing tools to correct generated content (GenArtist, ~\cite{wang2024genartist}) and reward-based fine-tuned LLMs (ComfyGen~\cite{gal2024comfygen}).
\begin{wrapfigure}[17]{r}{0.4\textwidth}
    \centering
    \includegraphics[width=0.4\textwidth]{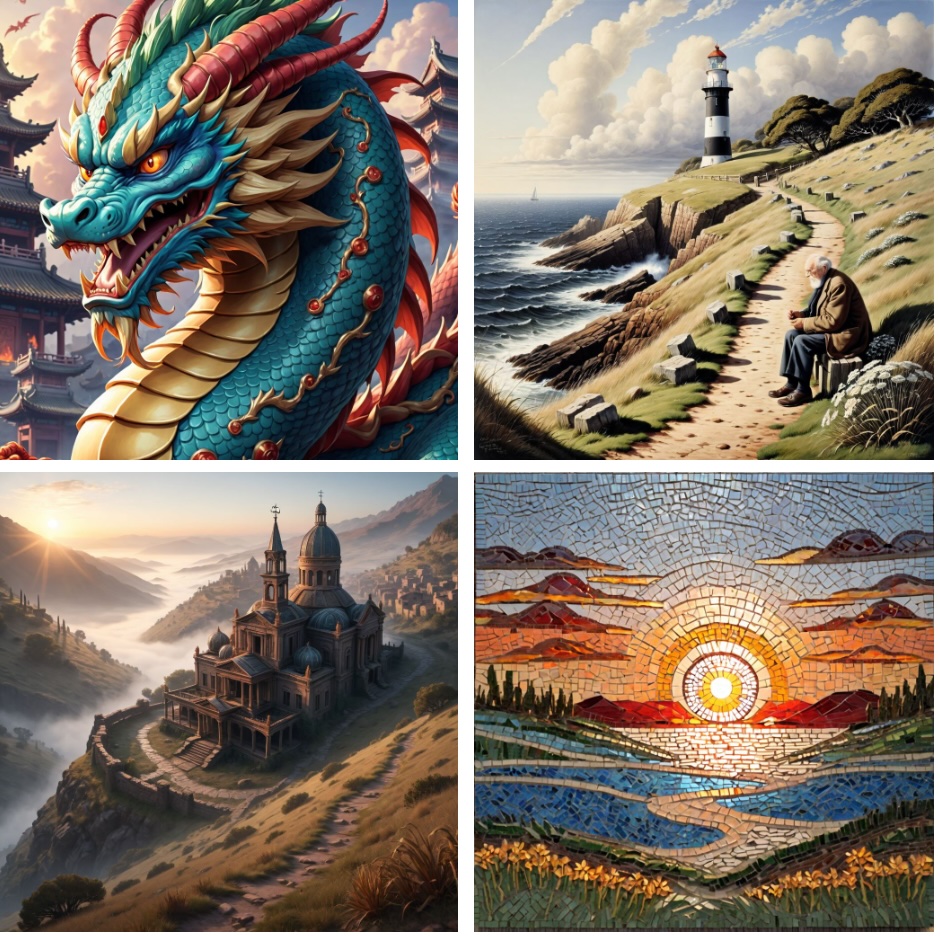}
    \caption{Example of generations with  \ourmethod{}}
    \label{fig:example}
\end{wrapfigure}
\paragraph{Prompt adherence: }
As summarized in Table~\ref{tab:combined_results},  \ourmethod{} demonstrates strong performance on the GenEval benchmark despite not being explicitly trained for prompt adherence. It achieves an overall score of $0.61$, matching the best-performing baseline, ComfyGen. Notably, our approach outperforms other methods in the “two objects” (0.85 vs. 0.82) and “binding” (0.38 vs. 0.29) categories, indicating improved capability in handling complex compositional prompts. A representative qualitative example illustrating prompt adherence is provided in Figure~\ref{fig:comparisons}.

\paragraph{Visual Quality:}
To automatically evaluate the visual quality of  \ourmethod{}'s outputs, we follow ~\cite{wallace2024diffusion,qi2024not,gal2024comfygen} and use a pair-wise comparison of HPS v2~\cite{wu2023human} score between \ourmethod{} and each baseline and report the average win rate. These comparisons use the full CivitAI test set of~\cite{gal2024comfygen}. The win-rate of each baseline over \ourmethod{} is reported in Table~\ref{tab:combined_results}. Additionally, we conducted a user study were we show users 35 randomly sampled prompts and the images generated for each, using \ourmethod{} and one of the baselines. Here, we focus on the best performing baseline from each category, as well as ComfyGen~\cite{gal2024comfygen}. We then ask them to select the image that they prefer, taking both prompt adherence and visual quality into account. We report the aggregated win percentage in figure~\ref{fig:user_study}, and add more details in the supplementary. 
This experiment demonstrates \ourmethod{}’s capability to create more performant ComfyUI workflows for the given input prompts. Representative qualitative comparisons highlighting these improvements are provided in Figure~\ref{fig:comparisons}, where our outputs consistently exhibit better prompt alignment and structural coherence compared to baseline generations.


\begin{table}[h]
    \fontsize{7.4pt}{7.4pt}\selectfont
    \centering
    \setlength{\tabcolsep}{1.4pt}
    \setlength{\belowcaptionskip}{-6pt}
    \begin{tabular}{lccccccc|c}
        \toprule
 \multirow{2}{*}{Model} & Single & Two & \multirow{2}{*}{Counting} & \multirow{2}{*}{Colors} & \multirow{2}{*}{Position} & Attribute & \multirow{2}{*}{Overall} & HPSv2 winrate \\
 &  object & object &  &  &  & binding & & vs. \ourmethod{} \\
\cmidrule(lr){1-1}
\cmidrule(lr){2-2}
\cmidrule(lr){3-8}
\cmidrule(lr){9-9}
SDXL & 0.98 & 0.74 & 0.39 & 0.85 & \underline{0.15} & 0.23 & 0.55 & 2\% $\pm$ 0.6\% \\
JuggernautXL & \textbf{1.00} & 0.73 & 0.48 & \underline{0.89} & 0.11 & 0.19 & 0.57 & 5\%$\pm$ 1\% \\
DreamShaperXL & \underline{0.99} & 0.78 & 0.45 & 0.81 & \textbf{0.17} & 0.24 & 0.57 & 3\%$\pm$ 0.6\% \\
DPO-SDXL & \textbf{1.00} & 0.81 & 0.44 & \textbf{0.90} & \underline{0.15} & 0.23 & \underline{0.59} & 5\%$\pm$ 1\% \\
\midrule
Most Popular Flow & 0.95 & 0.38 & 0.26 & 0.77 & 0.06 & 0.12 & 0.42 & 13\%$\pm$ 1\% \\
2\textsuperscript{nd} Most Popular Flow & \textbf{1.00} & 0.65 & \textbf{0.56} & 0.86 & 0.13 & 0.34 & \underline{0.59} & 14\%$\pm$ 1\% \\ 
\midrule
GenArtist & 0.94 & 0.41 & 0.40 & 0.72 & \textbf{0.24} & 0.07 & 0.47 & 5\% $\pm$ 1\%\\
RPG-DiffusionMaster & \textbf{1.00} & 0.64 & 0.21 & \underline{0.89} & \underline{0.20} & \underline{0.35} & 0.55 & 3\%$\pm$ 0.8\% \\

ComfyGen & \underline{0.99} & \underline{0.82} & \underline{0.50} & \textbf{0.90} & 0.13 & 0.29 & \textbf{0.61} & 40\% $\pm$ 2\%\\
\midrule
\ourmethod{} (Ours) & \textbf{1.00} & \textbf{0.85} & 0.44 & 0.86 & 0.11 & \textbf{0.38} & \textbf{0.61} & - \\
\bottomrule
    \end{tabular}
    \vspace{+2pt}
    \caption{GenEval and HPS v2 comparisons. \ourmethod{} is on-par with ComfyGen on GenEval and outperforms all other baseline approaches in overall score. On human preference metrics, \ourmethod{} significantly outperforms prior methods. CIs are calculated as one standard deviation from the mean.}
    \label{tab:combined_results}
\end{table}


\begin{figure}[ht]
\setlength{\belowcaptionskip}{-6pt}
    \centering
    \includegraphics[width=0.95\linewidth]{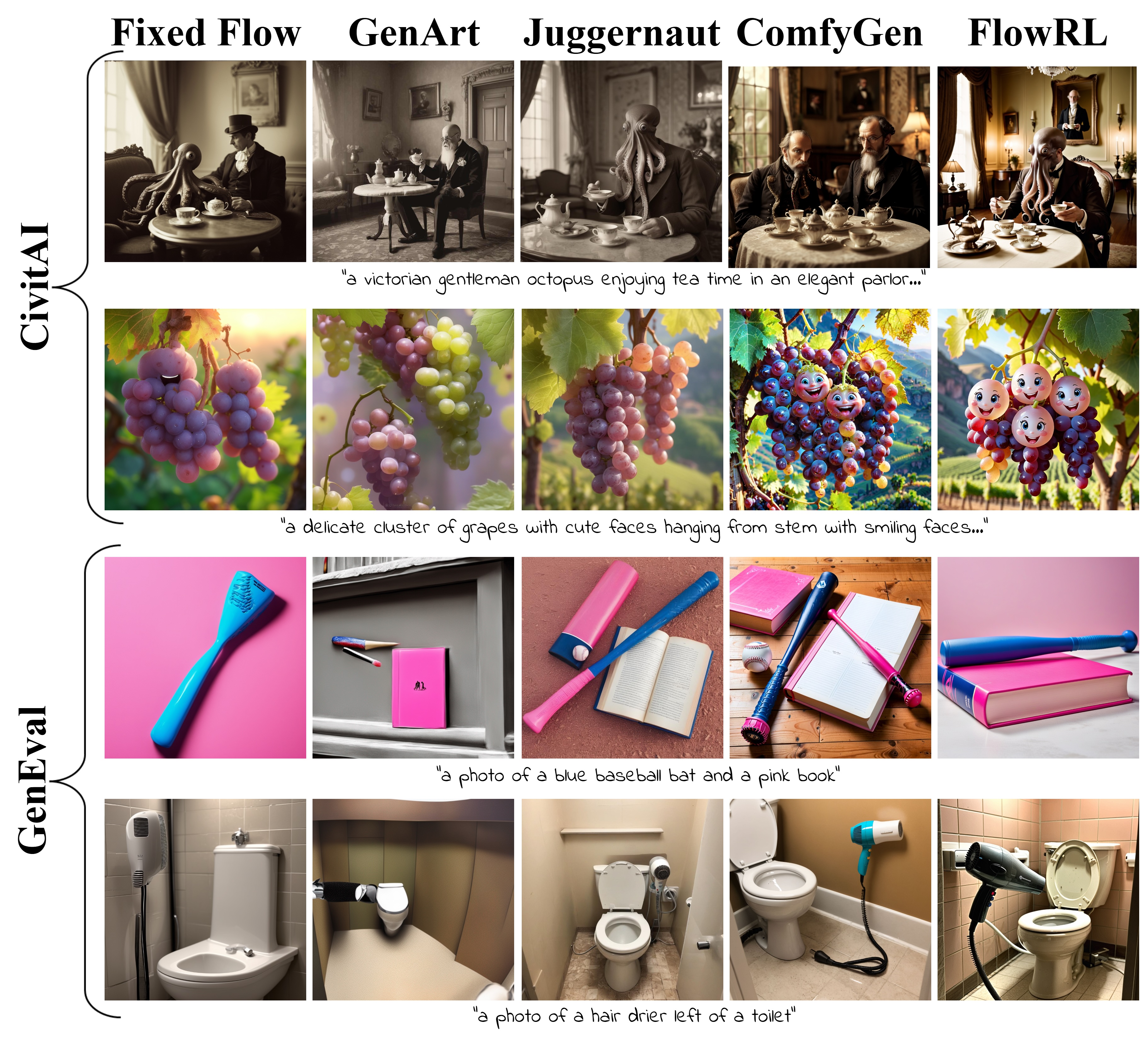}
    \caption{Qualitative results on CivitAI and GenEval prompts. }
    \label{fig:comparisons}
\end{figure}


\paragraph{Novelty of generated flows:}

A key advantage of our approach lies in its capacity to generate workflows that are not merely copies of those seen during training. 
To quantify this novelty, we generate 500 flows using the CivitAI test set, and calculate the normalized Levenshtein distance (NLD)~\cite{tashima2018fault,levenshtein1966} between each generated workflow and its nearest training sample. We further normalize these values by the NLD between training samples, giving us a measure of what fraction of the variance in training data we manage to preserve.
Additionally, we report how many generated flows exist ``as-is" in the training data, and how many unique flows were created in the $500$ output set.

\begin{figure}[t]
\setlength{\belowcaptionskip}{-6pt}
    \centering
    \includegraphics[width=0.7\textwidth]{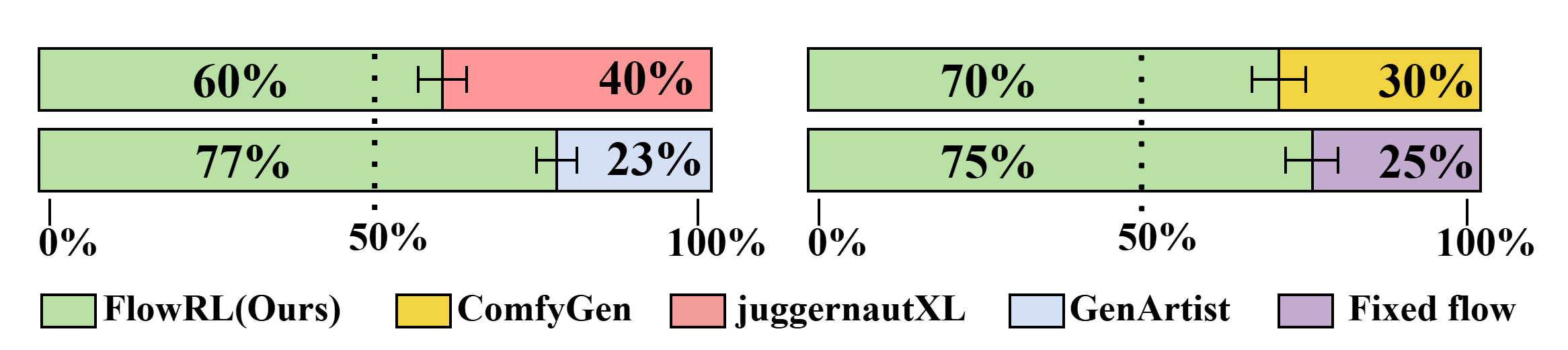}
    \caption{Human study win rate of  \ourmethod{} vs other relevant baselines}
    \label{fig:user_study}
\end{figure}
The results are reported in Table~\ref{tab:orig}. Our experiments confirm the findings of ~\cite{gal2024comfygen} which report that their approach learned to copy flows from the training data. \ourmethod{} meanwhile achieves significantly higher novelty, demonstrating the ability to generalize to new parameter combinations. These results highlight the effectiveness of our reinforcement learning framework in encouraging the LLM to explore and produce a broader range of complex workflows.


\paragraph{Effects of Dual model guidance:}
Next, we investigate the impact of our dual-model guidance approach. Specifically, prior work~\cite{dhariwal2021diffusionBeatsGAN} highlighted the ability of guidance-based methods to trade diversity for performance (or recall for precision). We show that similar behavior can be observed here. As shown in Table~\ref{tab:orig}, while increasing guidance strength ($\gamma$) improves HPS v2 scores win-rate vs ComfyGen, it significantly impacts the structural diversity of the generated workflows. At $\gamma=1.5$, our method maintains the uniqueness of generated flows. However, as $\gamma$ increases, we observe a dramatic reduction in the uniqueness ratio to just 8\%.

Notably, all our \ourmethod{} variants maintain near-zero overlap with training data (0-1\% "exists in data" vs ComfyGen's 94\%), and the NLD ratio actually improves with guidance (from 0.6 without CFG to 0.75 at $\gamma=2$). This pattern suggests that stronger guidance pushes the model to consistently generate a smaller subset of high-performing workflows, effectively concentrating probability mass on patterns that maximize reward but reducing exploration of the solution space. Conceptually, this mirrors observations in image generation with CFG, where higher guidance strengths produce higher-quality but less diverse output.
\begin{table}[ht]
\setlength{\belowcaptionskip}{-8pt}
    \fontsize{7.4pt}{7.4pt}\selectfont
\centering
\begin{tabular}{@{}lcccc@{}}
\toprule
\multirow{3}{*}{Method} & unique & exists& NLD& HPSv2 \\
& ratio & in data& ratio&win-rate\\
& (\%) & (\%) & &Vs ComfyGen\\
\midrule
ComfyGEN & 7\% & 94\% &0&-\\
\ourmethod{} (w/o CFG) & 41\% & 1\% &0.6&59\%\\
\ourmethod{} + CFG($\gamma=1.5$)& 41\% & 0\%  &0.74&60\%\\
\ourmethod{} + CFG ($\gamma=2$) & 8\% & 0\% &0.75&63\%\\
\bottomrule
\end{tabular}
\vspace{2pt}
\caption{Comparison of originality of flow generation models}
\vspace{-10pt}
\label{tab:orig}
\end{table}
\subsection{Ablation study}
To quantify the impact of individual components in  \ourmethod{}, we conducted an ablation study comparing variants with and without our key improvements. We evaluated the following modifications: (1) removing the component-aware reward model, (2) removing the uncertainty ensemble cutoff, (3) varying number of BERT models in our reward ensemble, (4) dropping the SFT step (stage 1), and (5) dropping the GRPO-tuning step. For (5), we instead use the stage-1 model to sample five flows per prompt, and use our reward ensemble to score them in relation to the prompt. Then, we generate an image with the highest scoring flow. Finally, to ensure that our benefits are not grounded in the novel encoding scheme, we also evaluate a baseline ComfyGen~\cite{gal2024comfygen} model trained on this new representation. We compare all scenarios against both the original ComfyGEN and against our full model, using HPSv2 scores on the CivitAI prompt set. The errors reported are the $1-\sigma$ Wald interval.

\begin{table}[ht]
\setlength{\belowcaptionskip}{-8pt}
\fontsize{7.4pt}{7.4pt}\selectfont
\centering
\begin{tabular}{@{}lccccccccc@{}}
\toprule
 & w/o  & w/o &  \multicolumn{3}{c}{Ensemble of} & & & w/o \\
\textbf{win ratio}&  prefix & reward  & 1 & 3 & 5 & ComfyGen  & SFT & SFT& \\
&  reward &  cutoff& \multicolumn{3}{c}{Berts}&(+ encoded)& only & stage\\
\midrule
\multirow{2}{*}{vs ComfyGen(\%)} & 55 &  57 & 55  & 56 & 56& 37& 29& 0 \\
 &$\pm$2.22 & $\pm$2.22 &$\pm$2.22 & $\pm$2.21& $\pm$2.22& $\pm$2.16 & $\pm$2.02& - \\
 \midrule
\multirow{2}{*}{vs ours (\%)} &  42 & 45  & 33 & 34 & 36&26 & 19& 0\\
&$\pm$2.21 & $\pm$2.21&$\pm$2.1 & $\pm$2.12& $\pm$2.15& $\pm$1.96 & $\pm$1.75& - \\
\bottomrule
\end{tabular}
\vspace{2pt}
\caption{The win ratio on the HPSv2 score for each component of our method compared to (1) the ComfyGen baseline and (2) the full ComfyGenRL model, using head-to-head comparisons.}
\label{tab:ablation}
\end{table}

The results are presented in table~\ref{tab:ablation}. These demonstrate the vital contribution of each component to overall performance. The full model consistently outperforms all ablations, with particularly significant drops observed when removing the SFT stage entirely (0\% win rate against ComfyGen and our full model). This emphasizes the critical nature of proper initialization before applying reinforcement learning methods. Looking at specific components, "prefix reward" proves the most beneficial, showing the importance of assigning more granular rewards. The "ComfyGen (+encoded)" variant, which uses our encoding scheme but lacks reinforcement learning, achieves only a 37\% win rate against the original ComfyGen, highlighting that our encoding improvements work synergistically with the GRPO training approach.


%% file: sections/5_discussion.tex
\section{Discussion}
\label{sec:discussion}
This paper presents a novel approach for fine-tuning LLMs using a combination of supervised learning on flow data, surrogate reward modeling, and uncertainty-aware reinforcement learning. Our method addresses several key challenges in LLM fine-tuning, including reward hacking, distribution shifts, and training efficiency.
The results demonstrate that our approach outperforms existing baselines across multiple metrics. Importantly, compared to prior workflow generation work, our approach demonstrates greater output diversity and successfully generalizes to novel flows that did not exist in the training data. 

Although it improves on the current state-of-the-art in multiple aspects, our approach still maintains many of their limitations. First, it remains focused on text-to-image workflows, with no support for editing tasks or video modules. Second, introducing new workflow components to the LLM would require retraining our entire stack. In the future, we hope to explore more efficient ways of adapting to novel models or blocks.



By enabling reliable and diverse automated workflow generation, our work advances generative AI systems that adapt to human preferences. We hope it will help foster more collaborative innovation by streamlining the integration of independently trained, specialized modules.


%% file: sections/jpg_sections/appendix.tex
\section{Appendix}
\subsection{Broader impact statement}

Our work offers a new path to improve text-to-image generation, but this improvement is not without possible social impacts. Text-to-image models can be used to create harmful or misleading content, and improving their output can increase this risk. 

Moreover, our work relies on the abundance of community-created, fine-tuned specialized models and adapters. These are rarely developed with safety in mind, and do not typically undergo red team assessments. Hence, they may increase the risk of the user generating biased or unsafe content. However, this can be mitigated by carefully curating the generative models seen during training, or by black-listing specific models in the output flow strings.

Future work may be able to further refine the reward model used at training to also align it with safety, for example by reducing the score for content deemed unsafe by a detector.

\subsection{ComfyUI Overview}
ComfyUI is a popular ($77,300$ stars on GitHub at the time of this writing), open-source workflow engine designed for flexible and extensible automation of generative AI tasks. Its node-based interface allows users to visually construct and execute complex processing pipelines, with users across the community often implementing and sharing new nodes to accommodate the changing landscape of generative tools. The pipelines constructed in ComfyUI can be exported to a JSON format, which we then map to a more compact representation and use for both our training data and our LLM output representation. To run our generated workflows, we convert them back to the JSON format and run them through the ComfyUI API. A dedicated user could also load these workflows through the UI and further manually refine them.

\subsection{Additional Qualitative results}
Here, we give more qualitative examples of \ourmethod{} generations.

\begin{figure}[ht]
    \centering
    \begin{subfigure}[b]{0.32\linewidth}
        \centering
        \includegraphics[width=\linewidth]{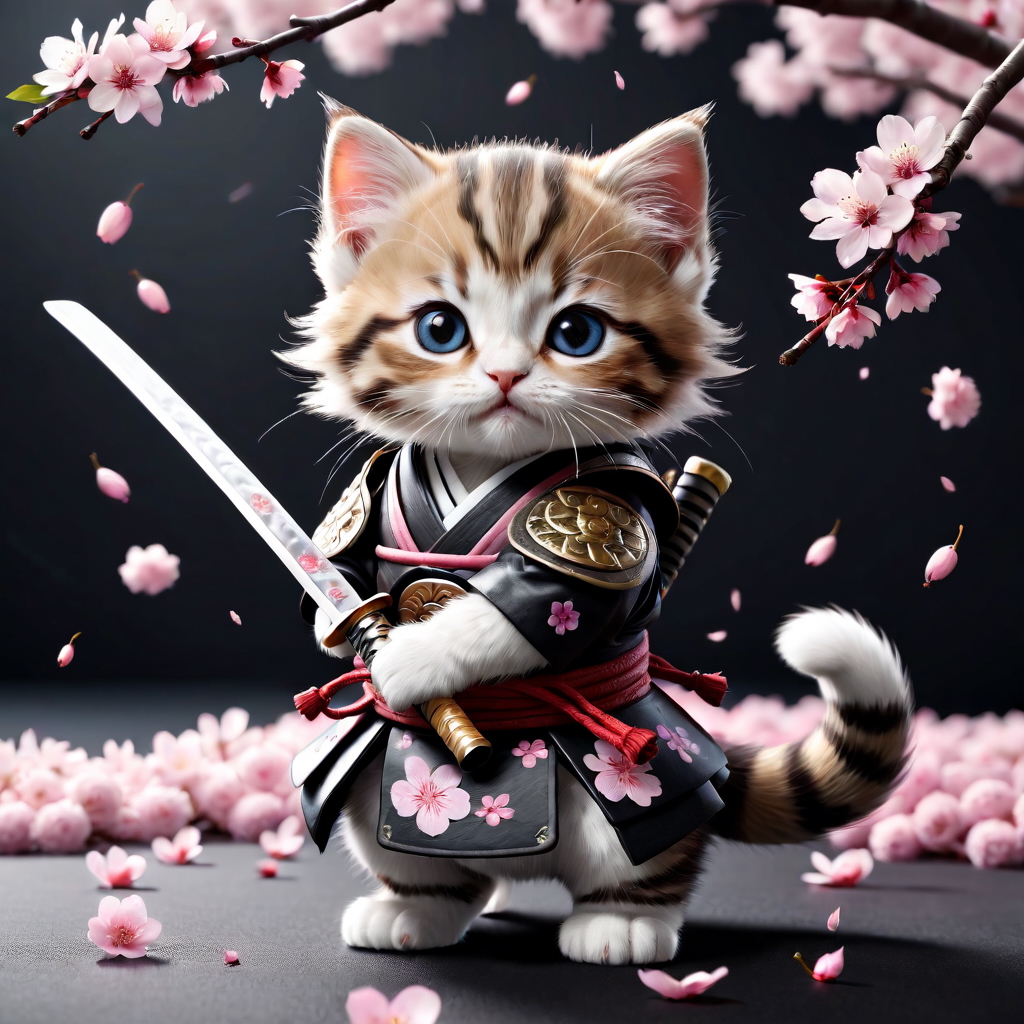}
    \end{subfigure}
    \hfill
    \begin{subfigure}[b]{0.32\linewidth}
        \centering
        \includegraphics[width=\linewidth]{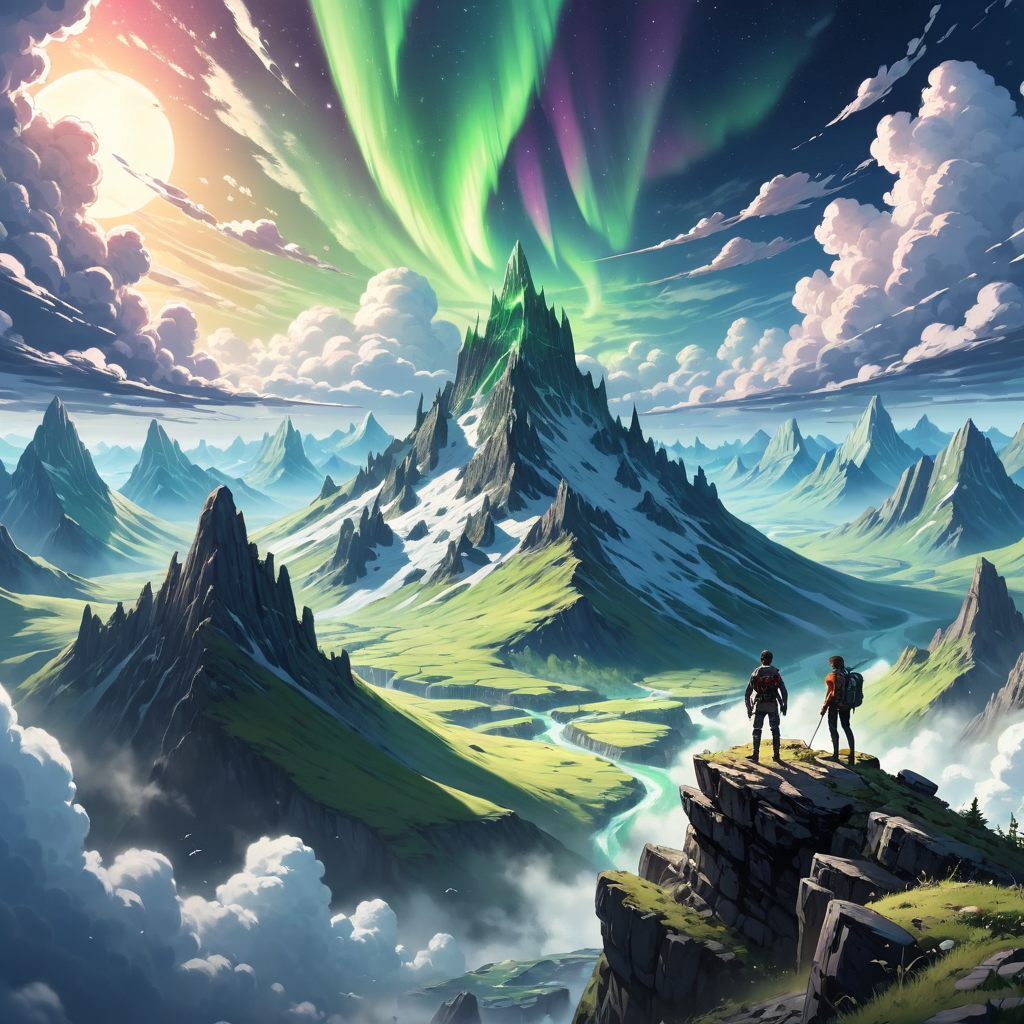}
    \end{subfigure}
    \hfill
    \begin{subfigure}[b]{0.32\linewidth}
        \centering
        \includegraphics[width=\linewidth]{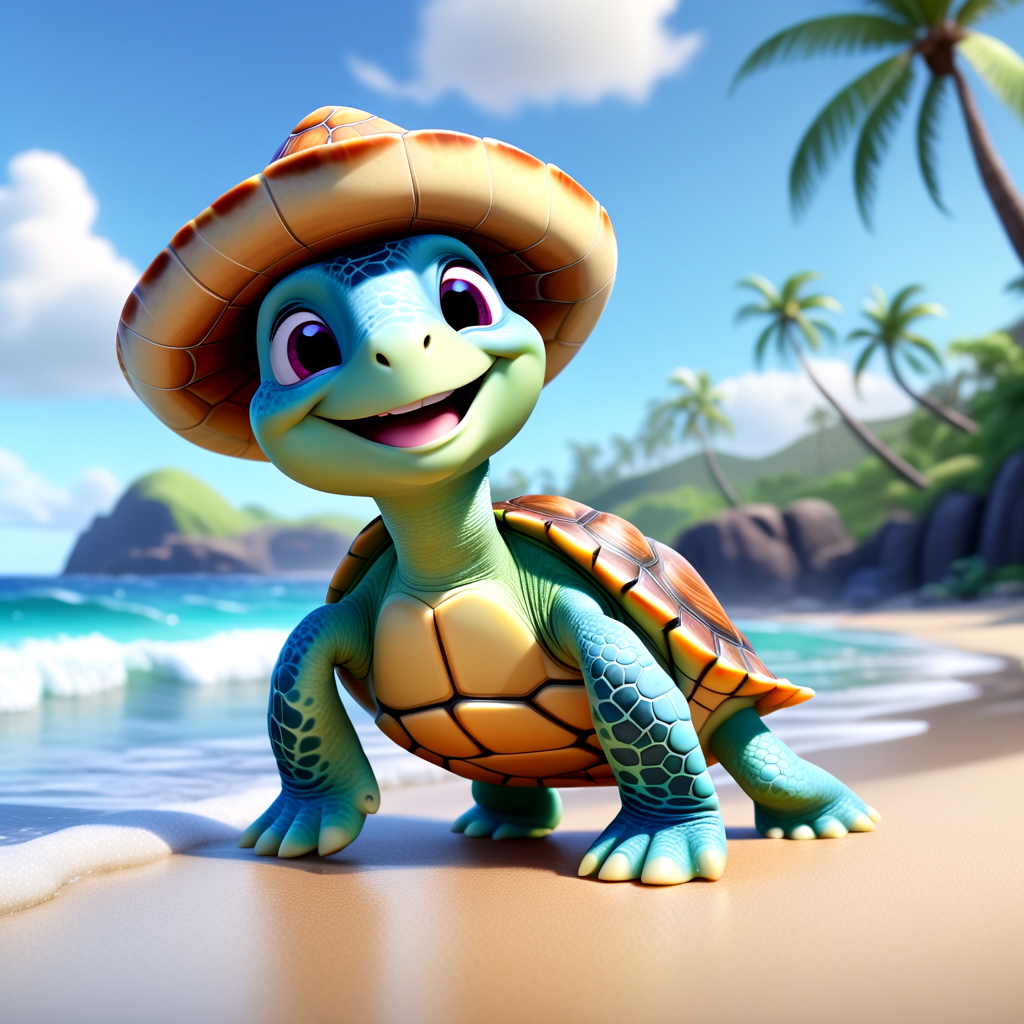}
    \end{subfigure}

    \vskip\baselineskip
    \begin{subfigure}[b]{0.32\linewidth}
        \centering
        \includegraphics[width=\linewidth]{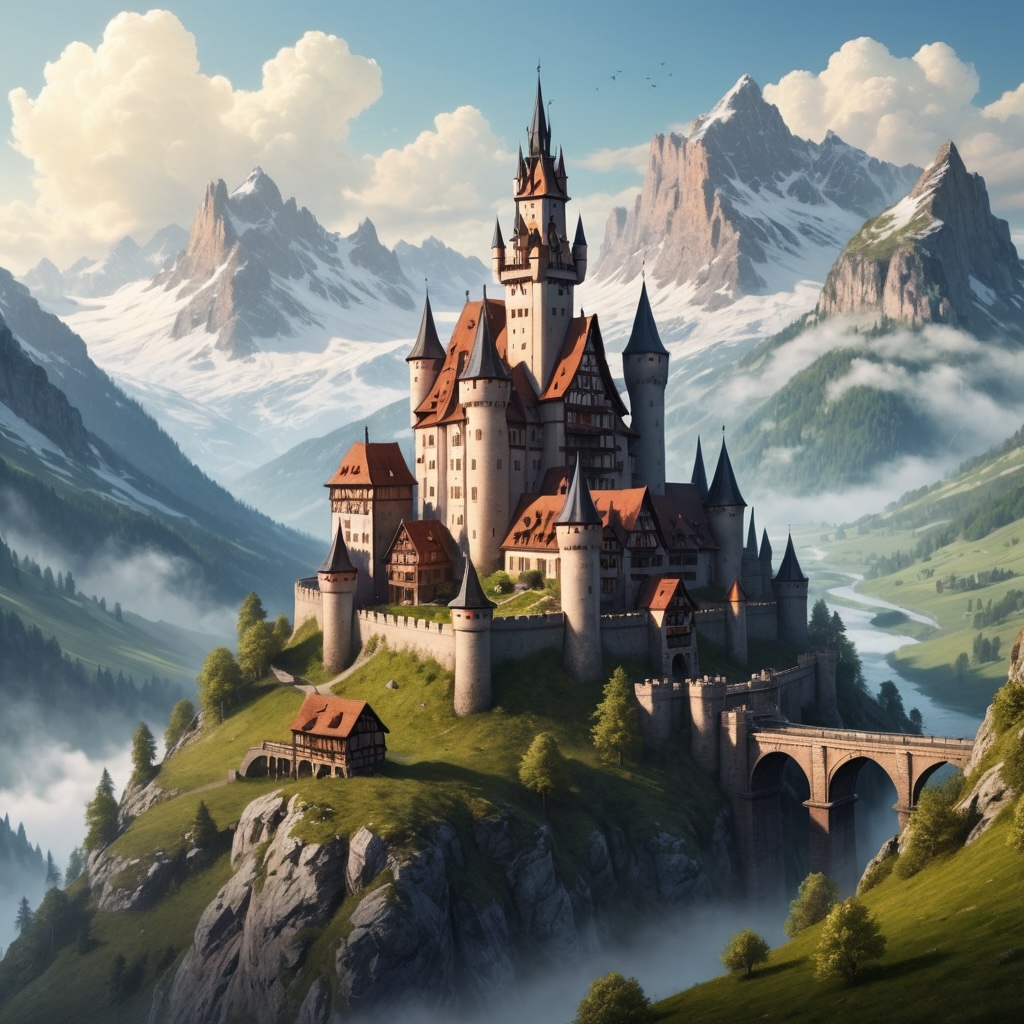}
    \end{subfigure}
    \hfill
    \begin{subfigure}[b]{0.32\linewidth}
        \centering
        \includegraphics[width=\linewidth]{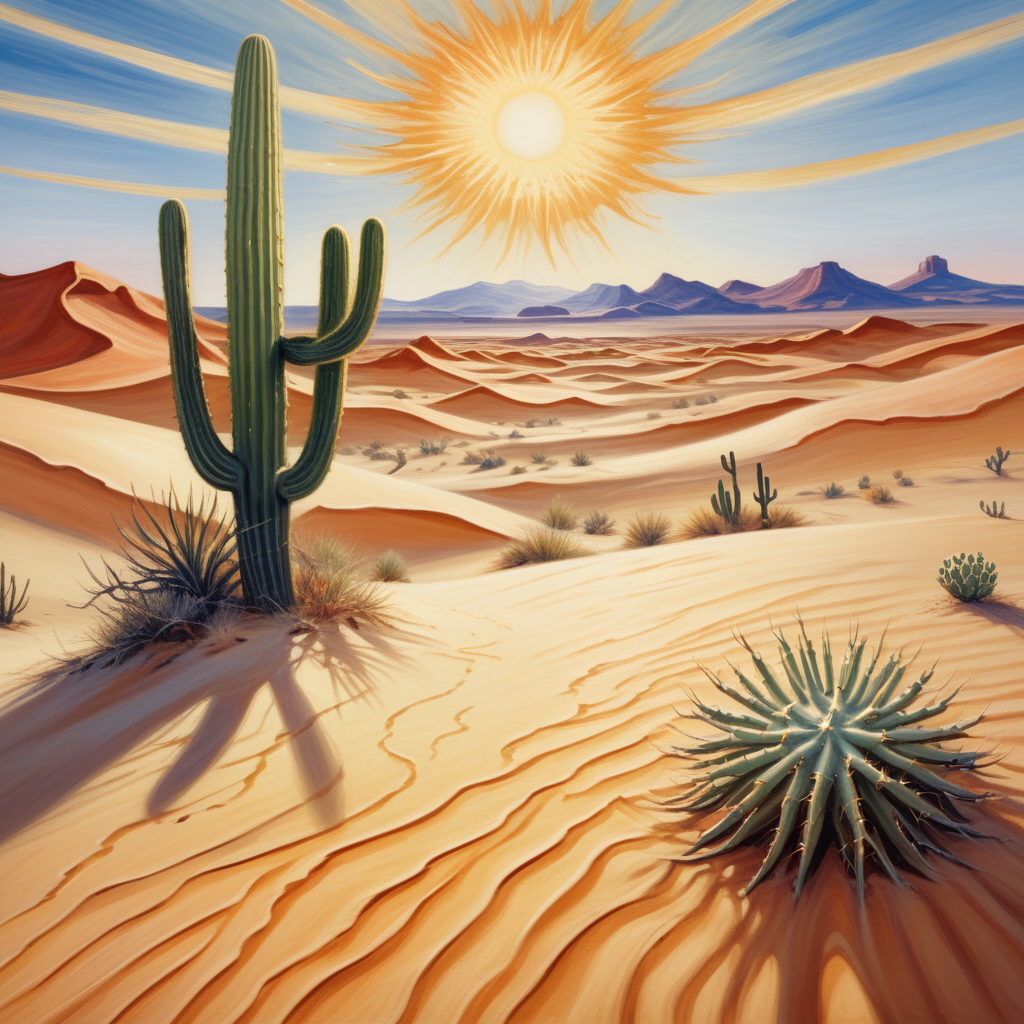}
    \end{subfigure}
    \hfill
    \begin{subfigure}[b]{0.32\linewidth}
        \centering
        \includegraphics[width=\linewidth]{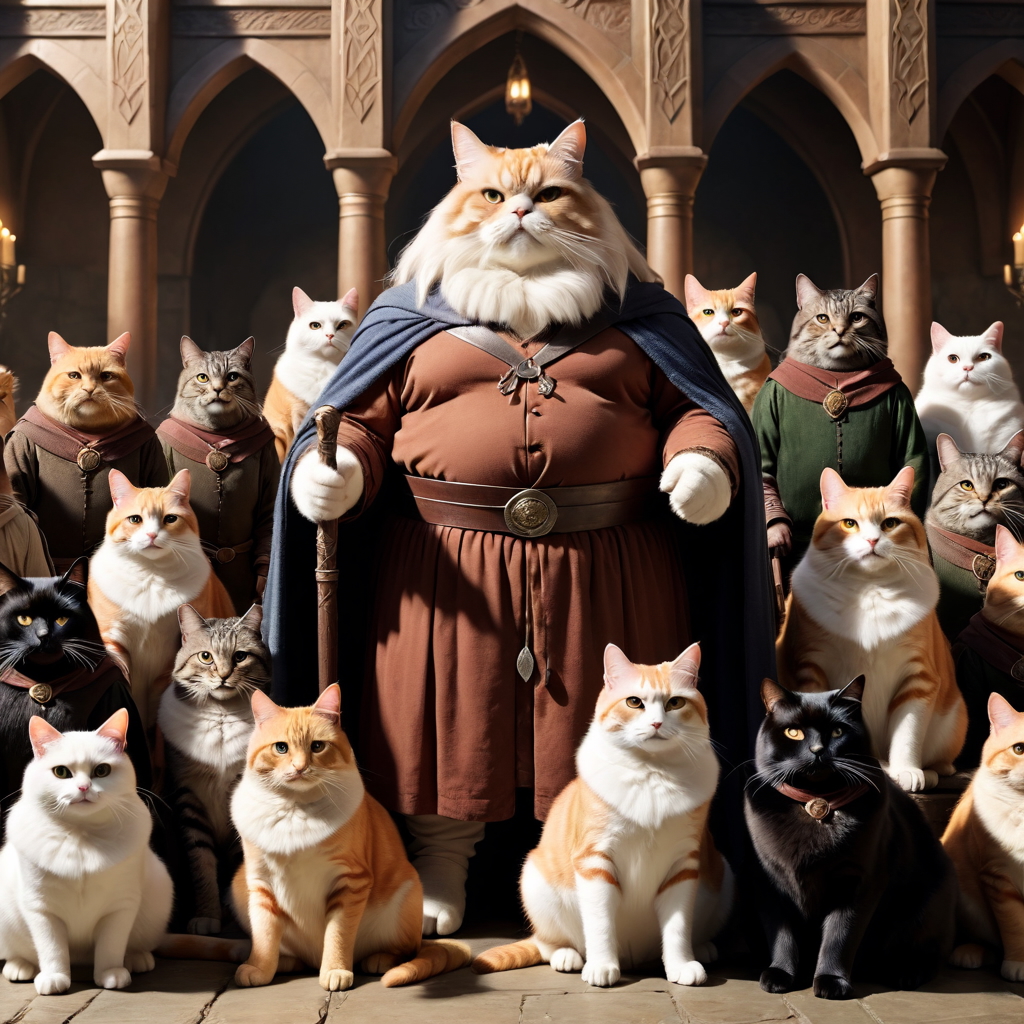}
    \end{subfigure}

    \vskip\baselineskip
    \begin{subfigure}[b]{0.32\linewidth}
        \centering
        \includegraphics[width=\linewidth]{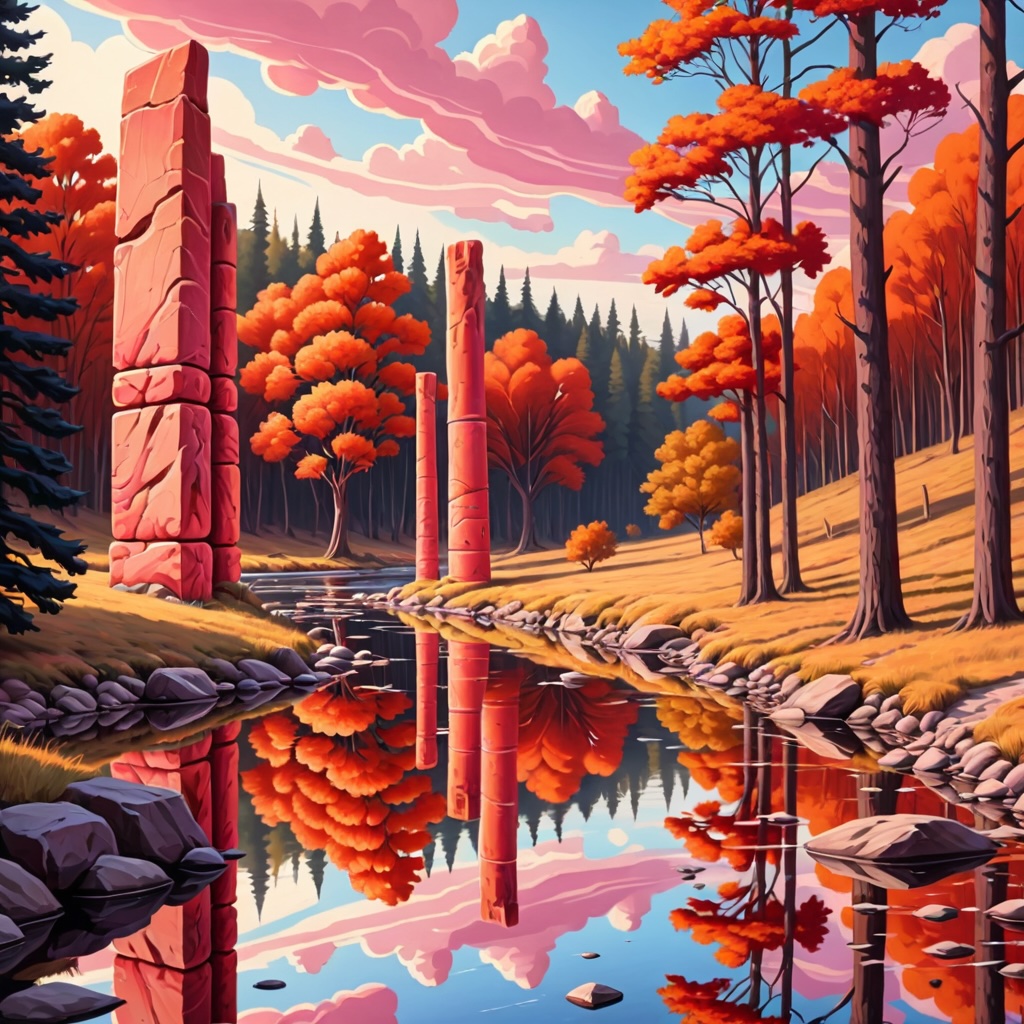}
    \end{subfigure}
    \hfill
    \begin{subfigure}[b]{0.32\linewidth}
        \centering
        \includegraphics[width=\linewidth]{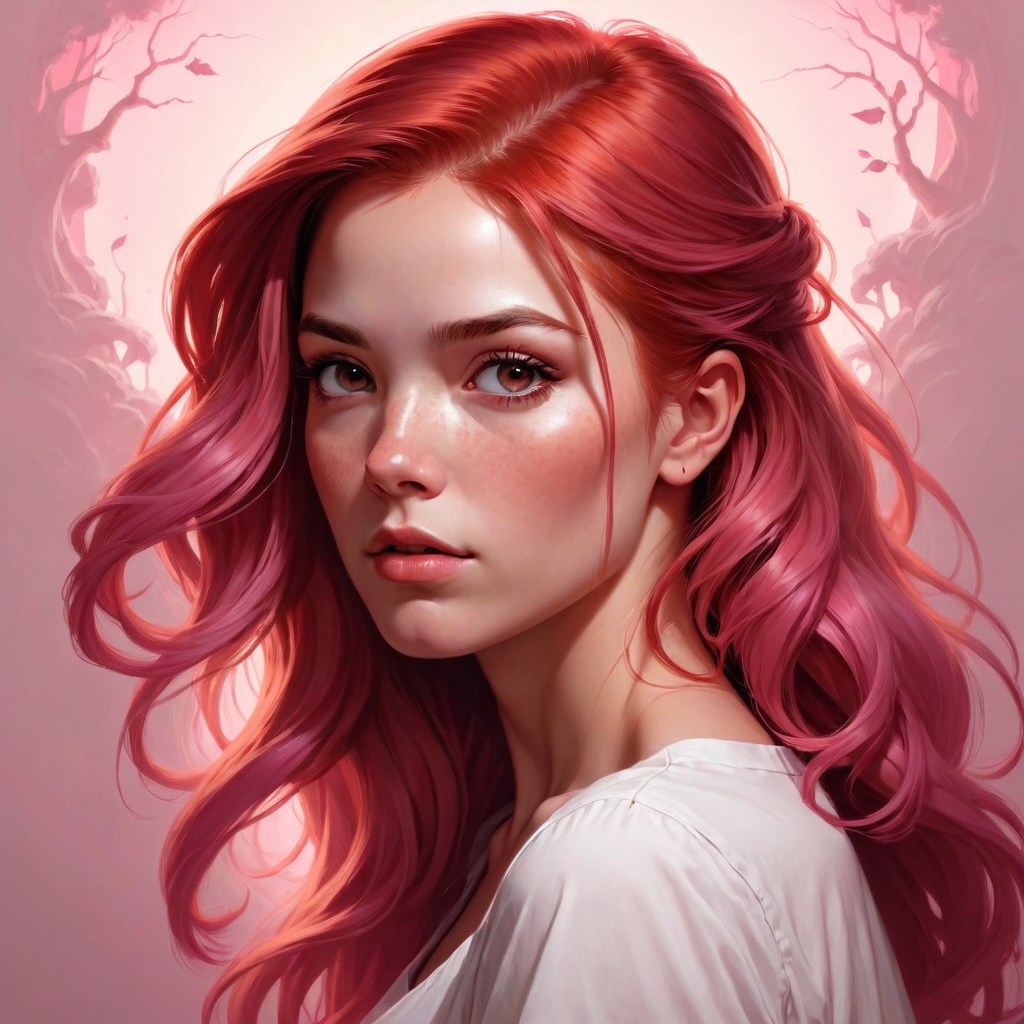}
    \end{subfigure}
    \hfill
    \begin{subfigure}[b]{0.32\linewidth}
        \centering
        \includegraphics[width=\linewidth]{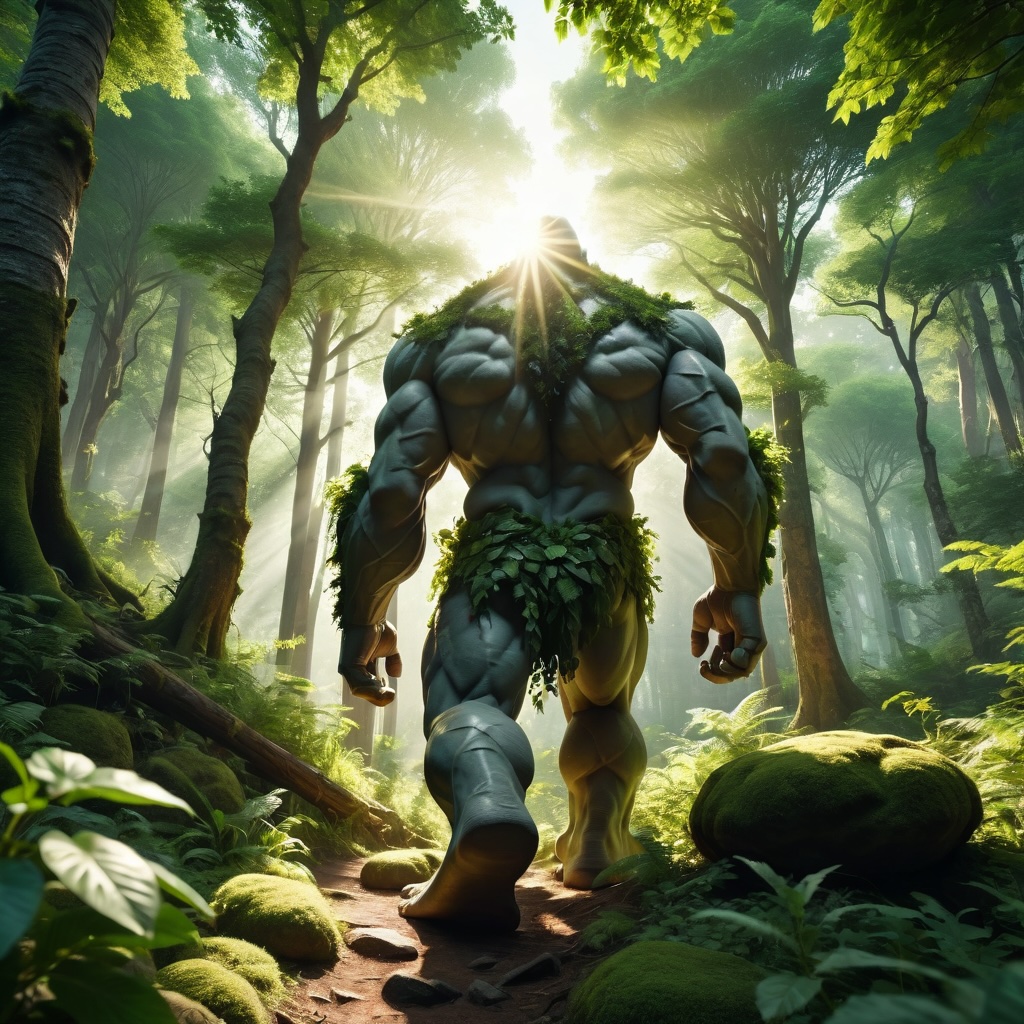}
    \end{subfigure}

    \vskip\baselineskip
    \begin{subfigure}[b]{0.32\linewidth}
        \centering
        \includegraphics[width=\linewidth]{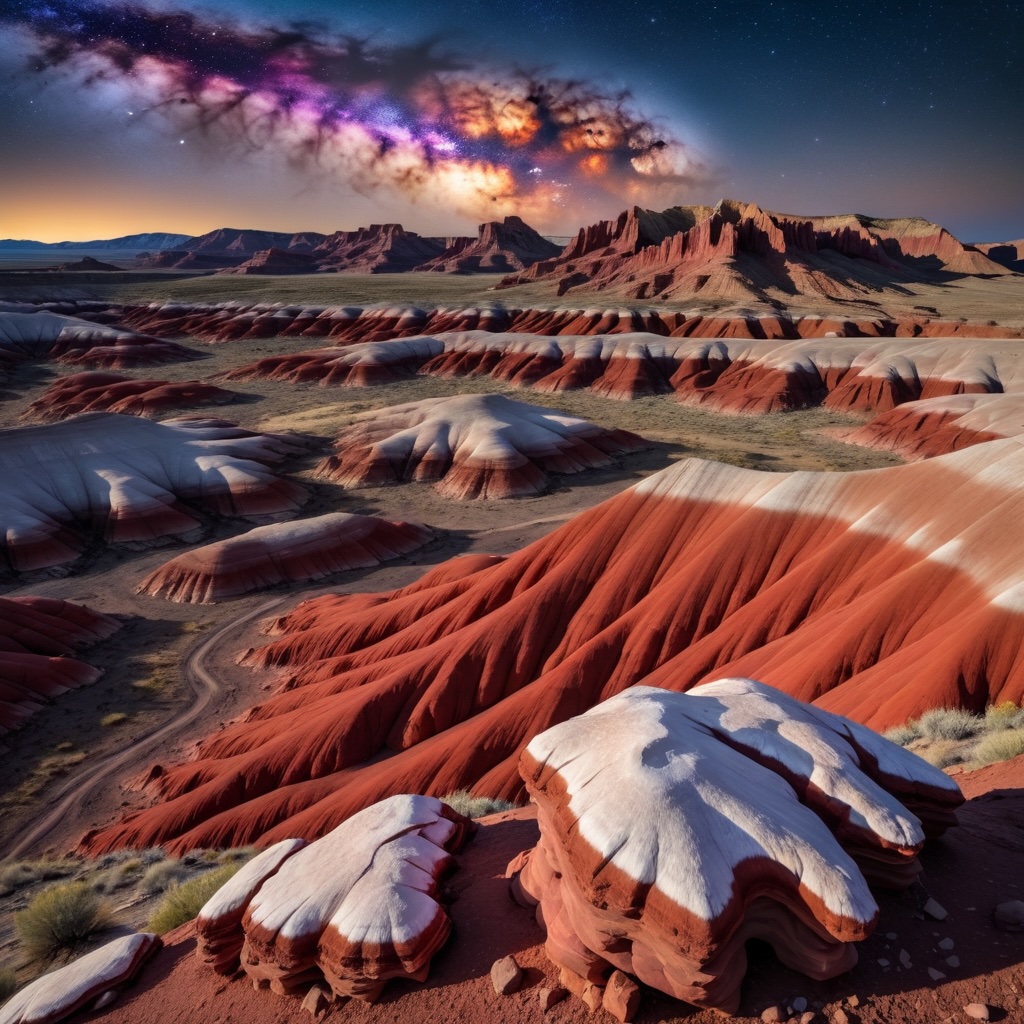}
    \end{subfigure}
    \hfill
    \begin{subfigure}[b]{0.32\linewidth}
        \centering
        \includegraphics[width=\linewidth]{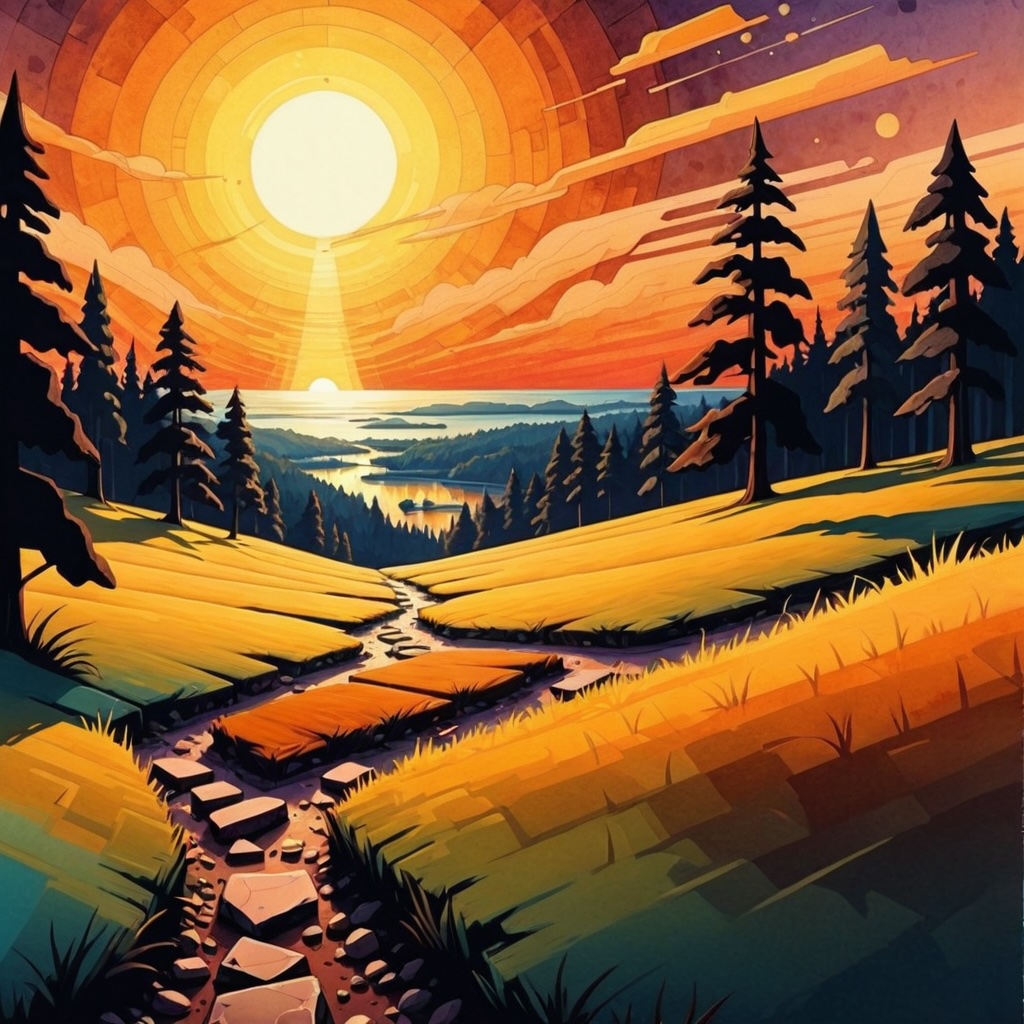}
    \end{subfigure}
    \hfill
    \begin{subfigure}[b]{0.32\linewidth}
        \centering
        \includegraphics[width=\linewidth]{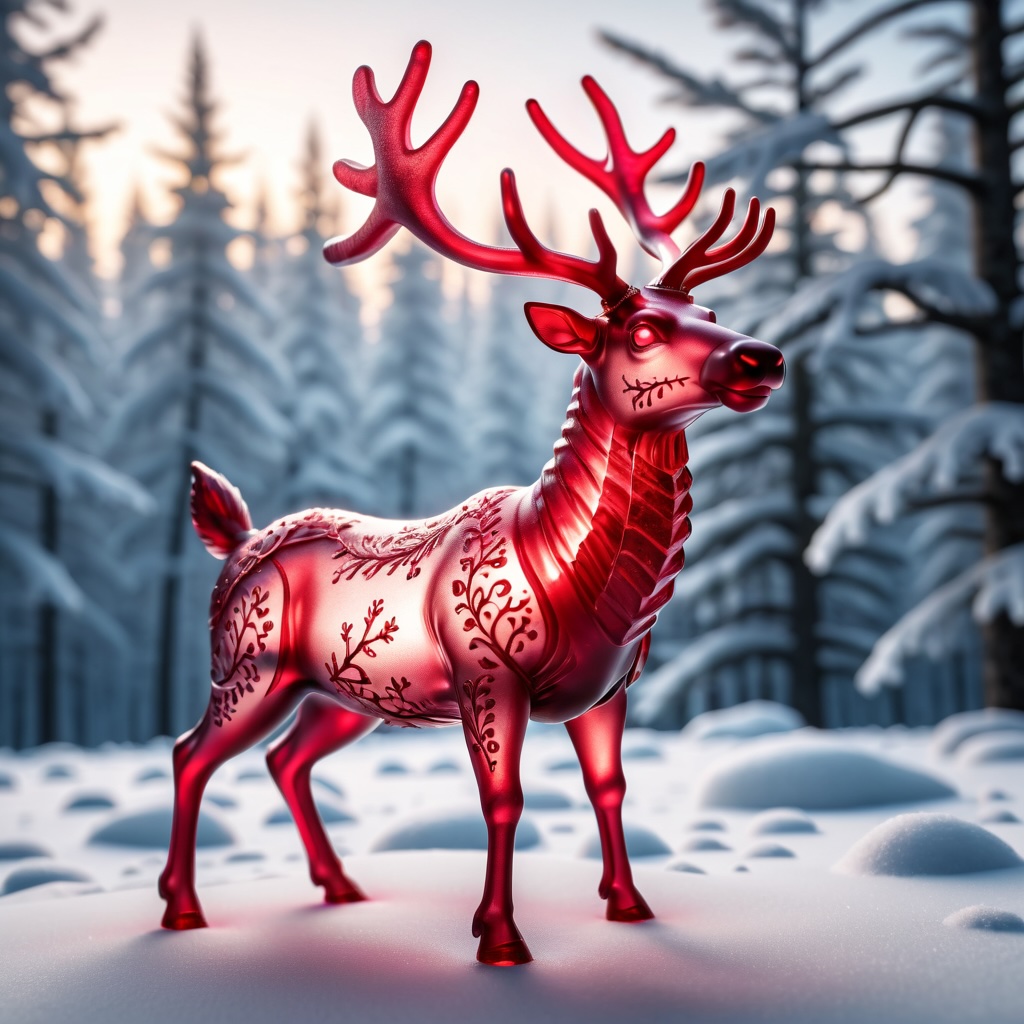}
    \end{subfigure}
\caption{More qualitative generations using \ourmethod{}}
    \label{fig:generations_3}
\end{figure}

\begin{figure}[ht]
    \centering
    \begin{subfigure}[b]{0.32\linewidth}
        \centering
        \includegraphics[width=\linewidth]{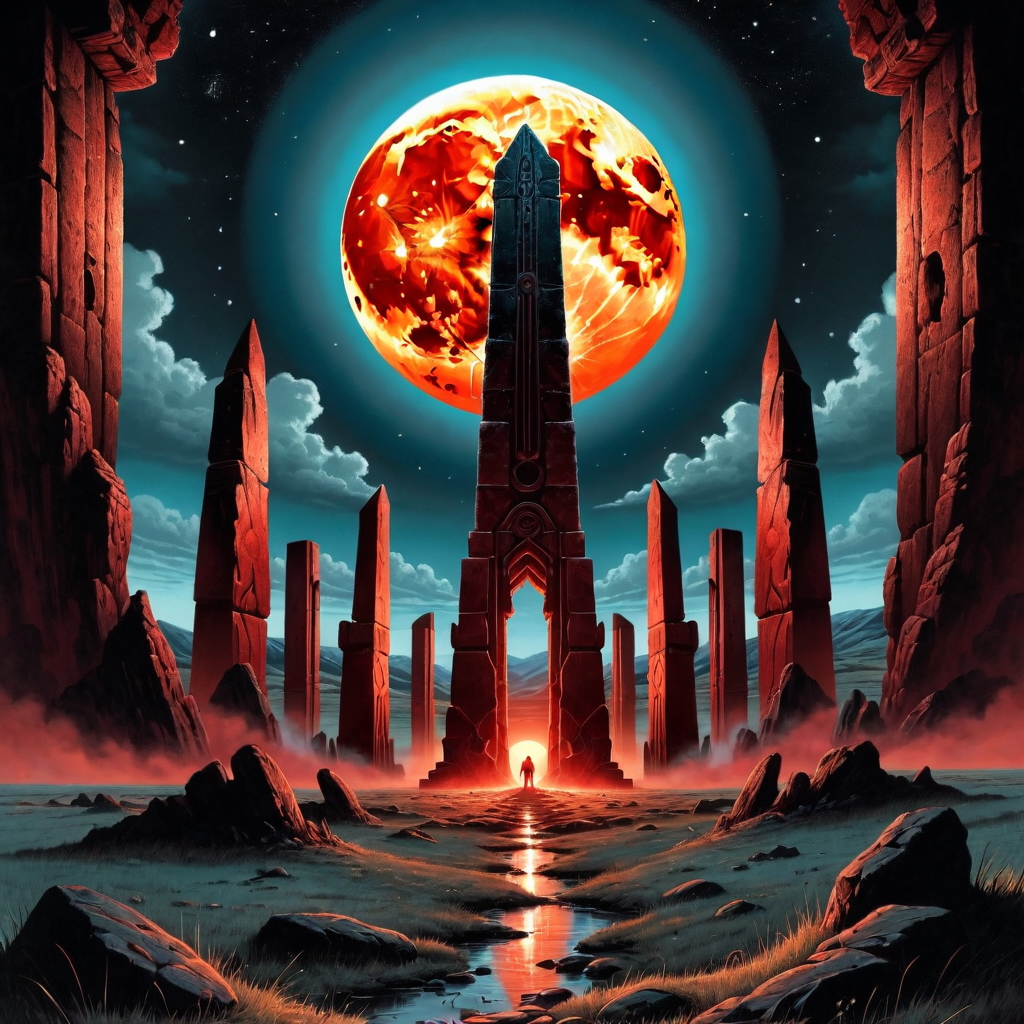}
    \end{subfigure}
    \hfill
    \begin{subfigure}[b]{0.32\linewidth}
        \centering
        \includegraphics[width=\linewidth]{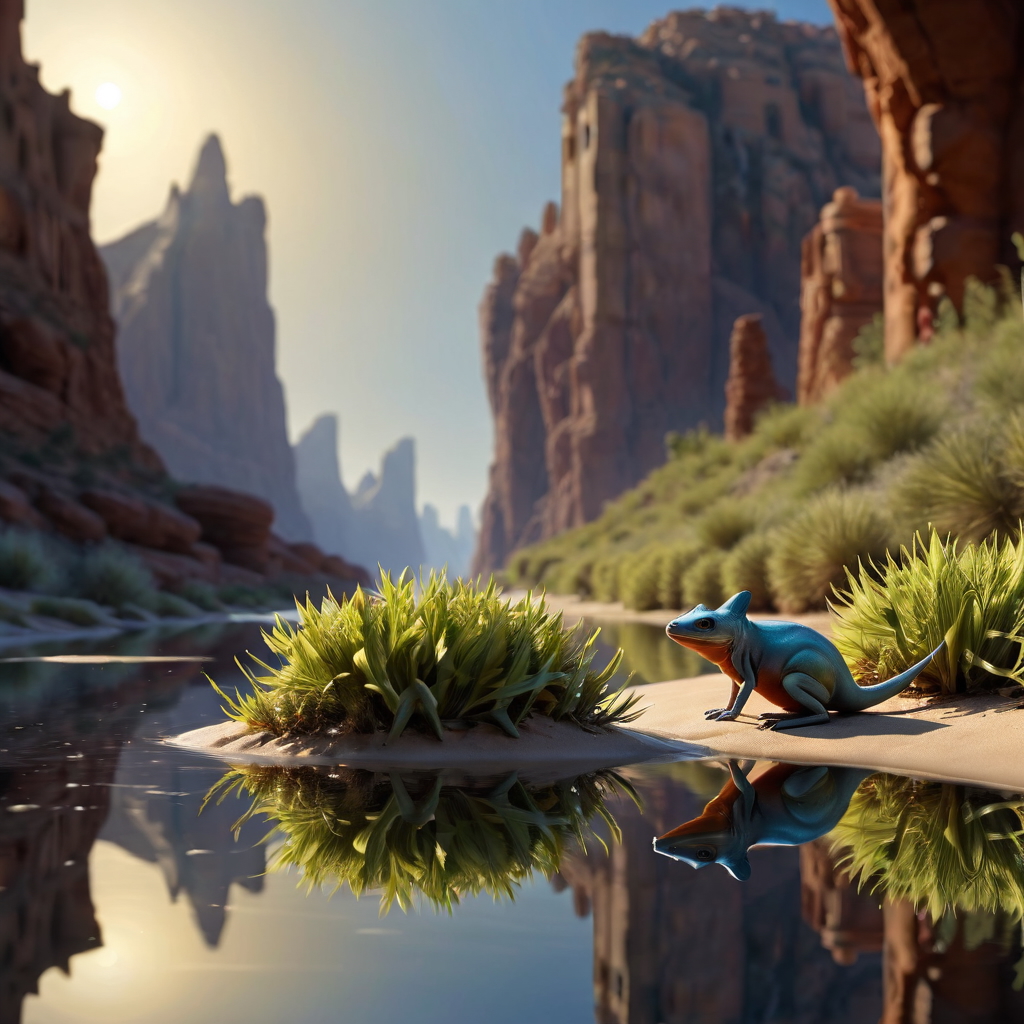}
    \end{subfigure}
    \hfill
    \begin{subfigure}[b]{0.32\linewidth}
        \centering
        \includegraphics[width=\linewidth]{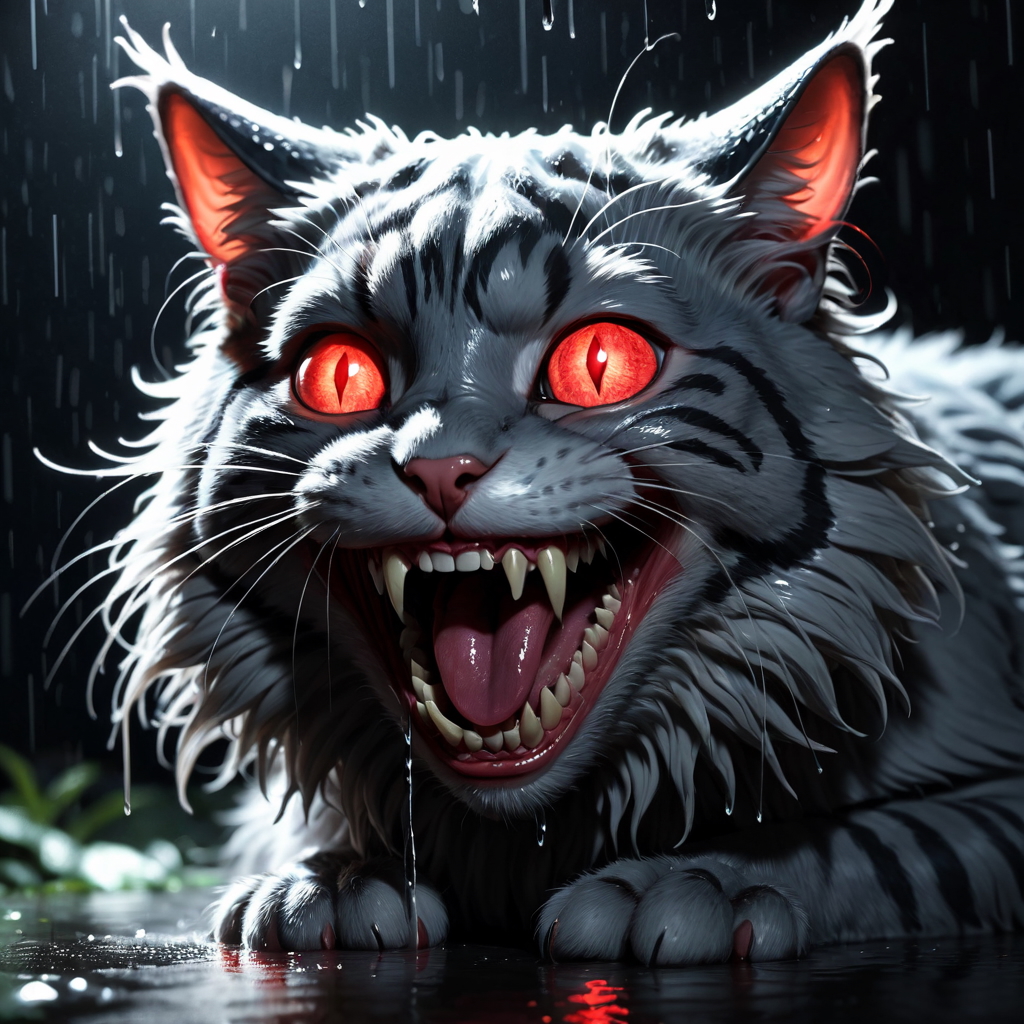}
    \end{subfigure}

    \vskip\baselineskip
    \begin{subfigure}[b]{0.32\linewidth}
        \centering
        \includegraphics[width=\linewidth]{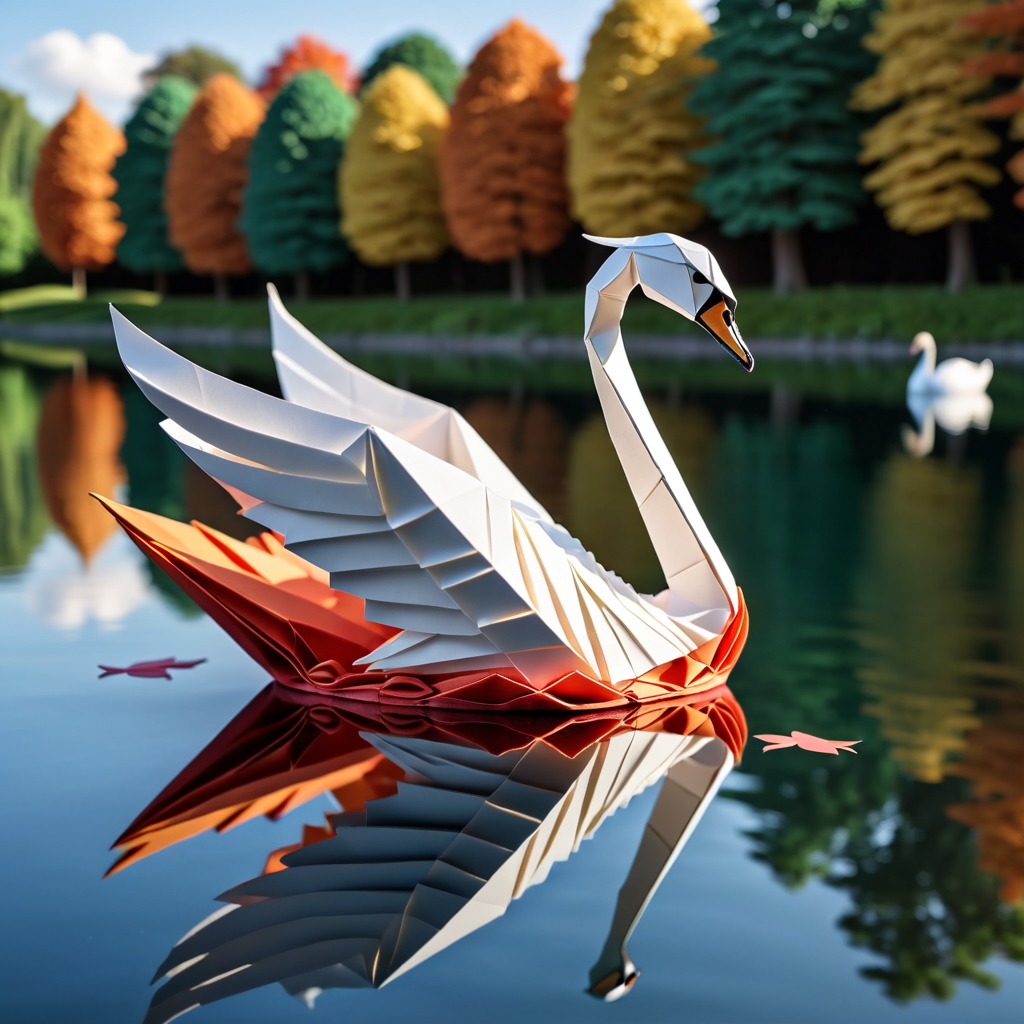}
    \end{subfigure}
    \hfill
    \begin{subfigure}[b]{0.32\linewidth}
        \centering
        \includegraphics[width=\linewidth]{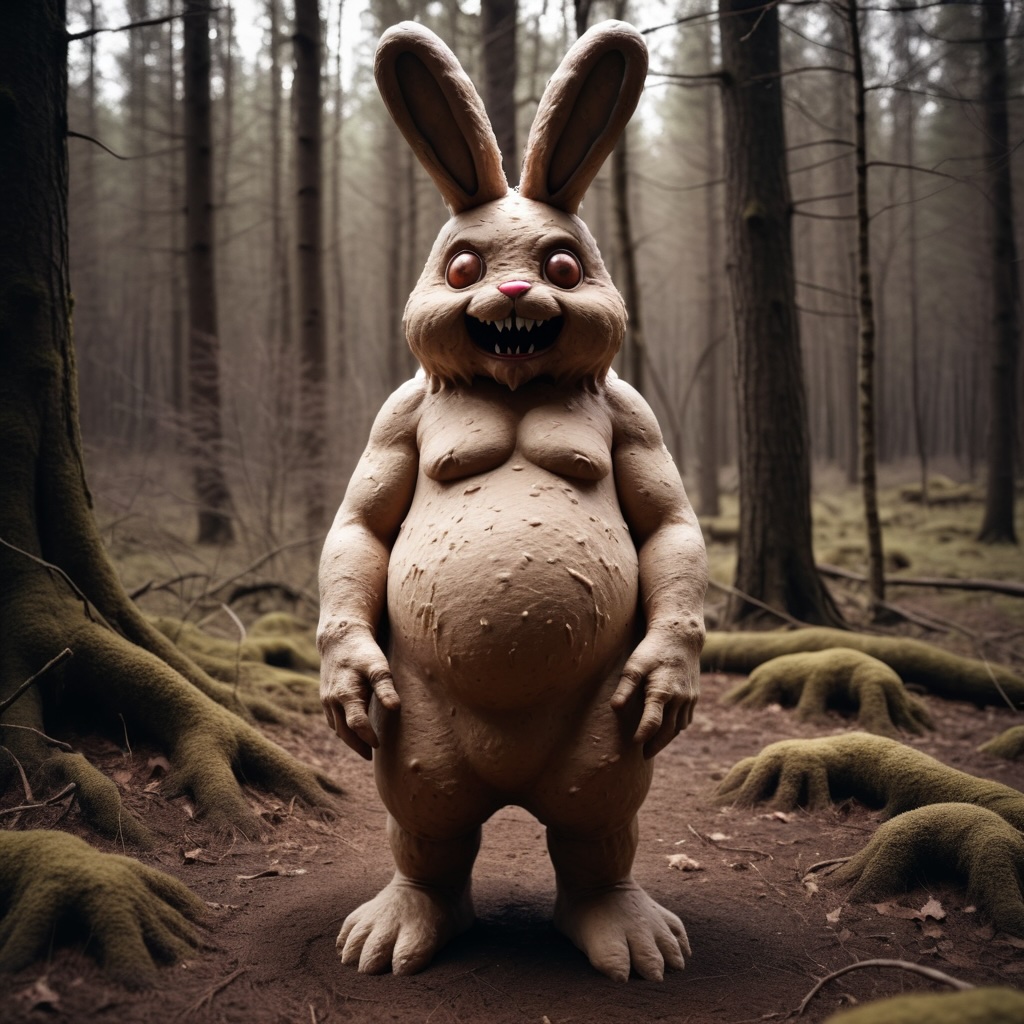}
    \end{subfigure}
    \hfill
    \begin{subfigure}[b]{0.32\linewidth}
        \centering
        \includegraphics[width=\linewidth]{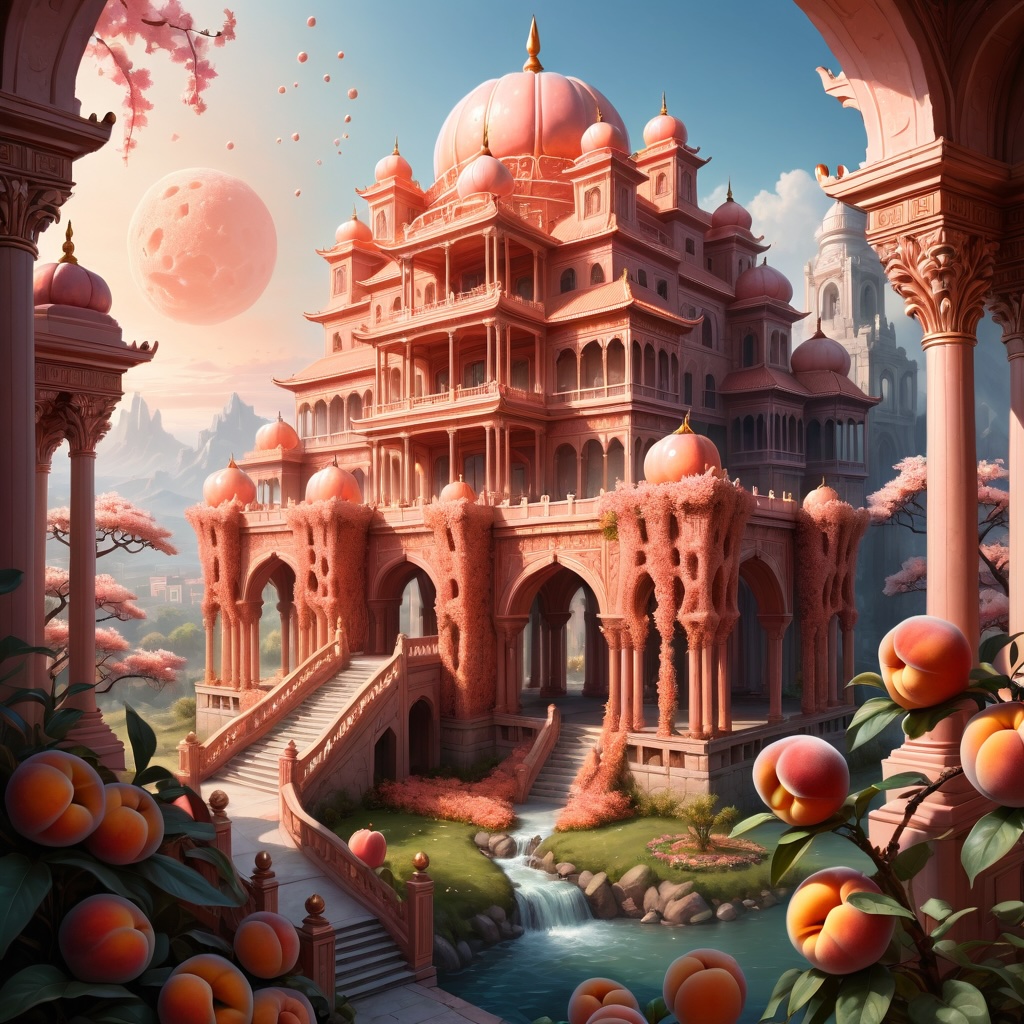}
    \end{subfigure}

    \vskip\baselineskip
    \begin{subfigure}[b]{0.32\linewidth}
        \centering
        \includegraphics[width=\linewidth]{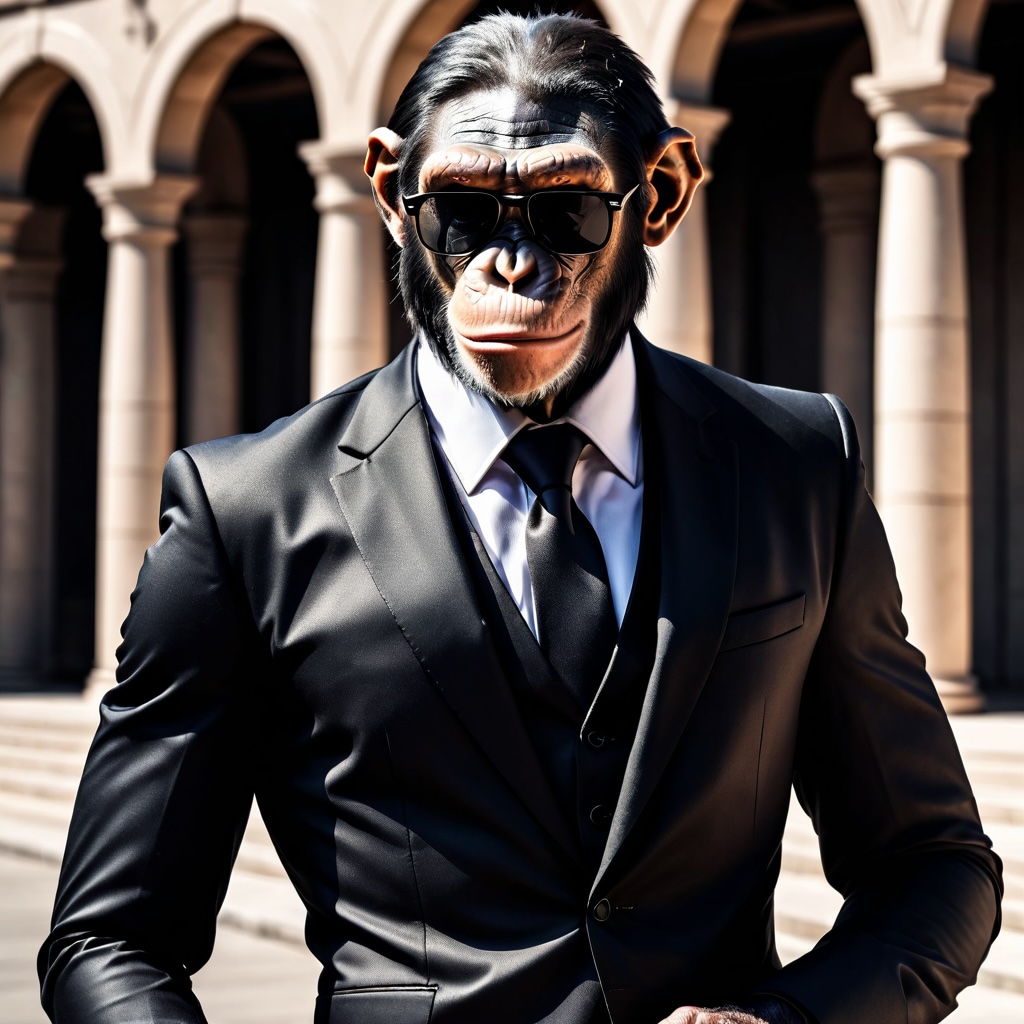}
    \end{subfigure}
    \hfill
    \begin{subfigure}[b]{0.32\linewidth}
        \centering
        \includegraphics[width=\linewidth]{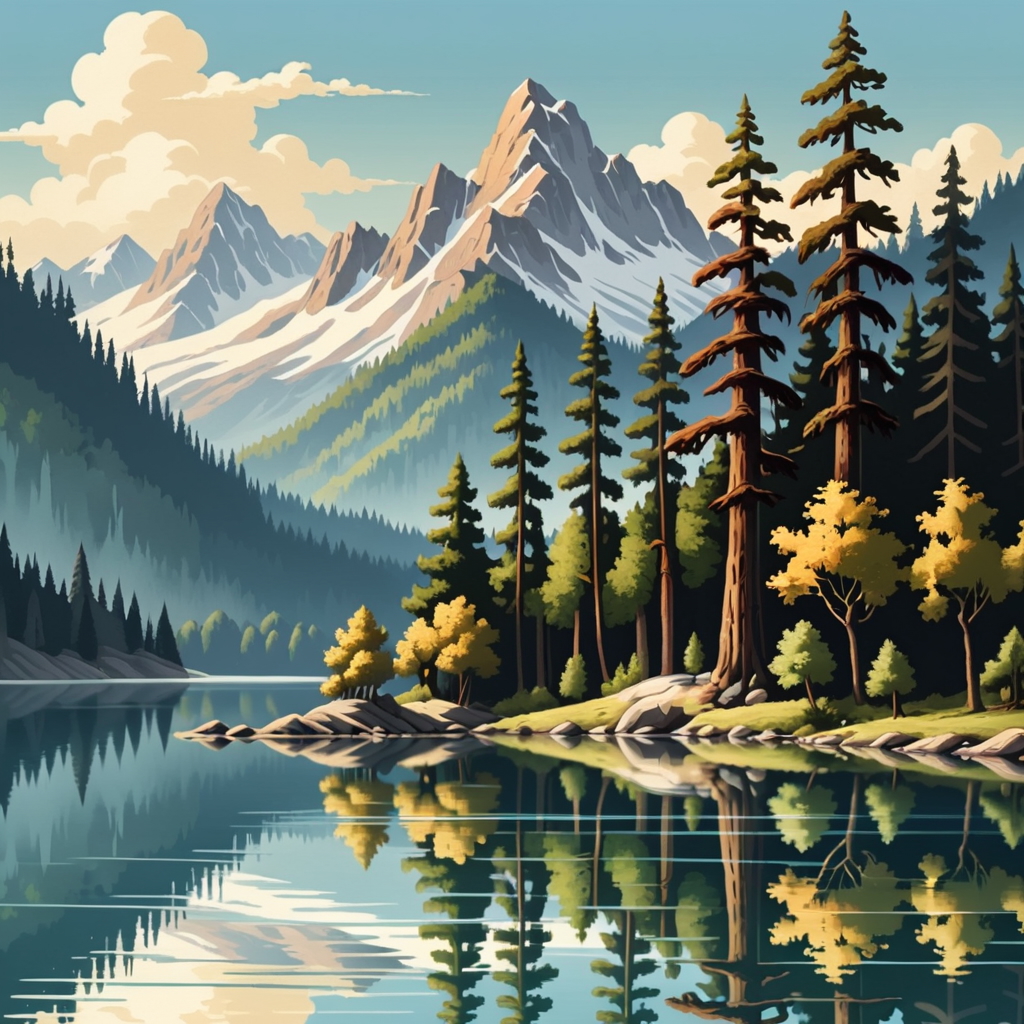}
    \end{subfigure}
    \hfill
    \begin{subfigure}[b]{0.32\linewidth}
        \centering
        \includegraphics[width=\linewidth]{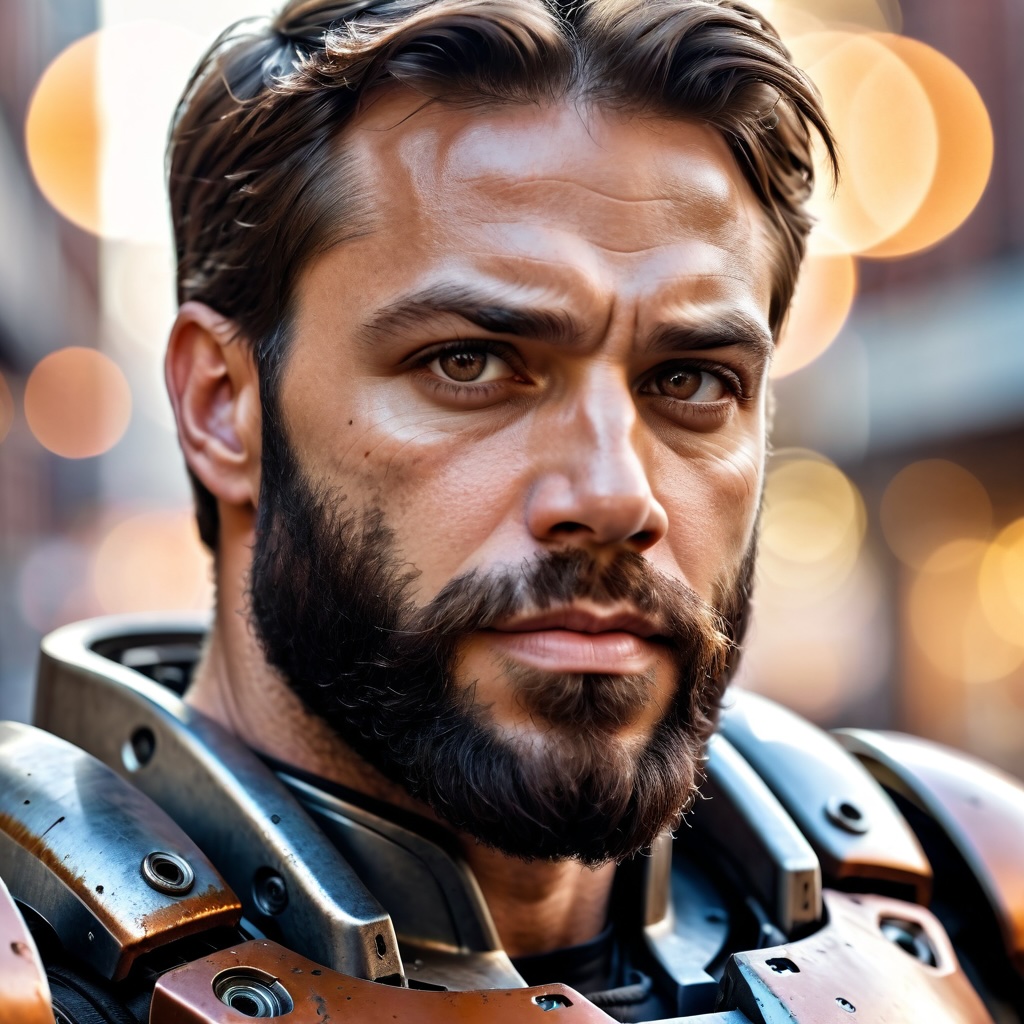}
    \end{subfigure}

    \vskip\baselineskip
    \begin{subfigure}[b]{0.32\linewidth}
        \centering
        \includegraphics[width=\linewidth]{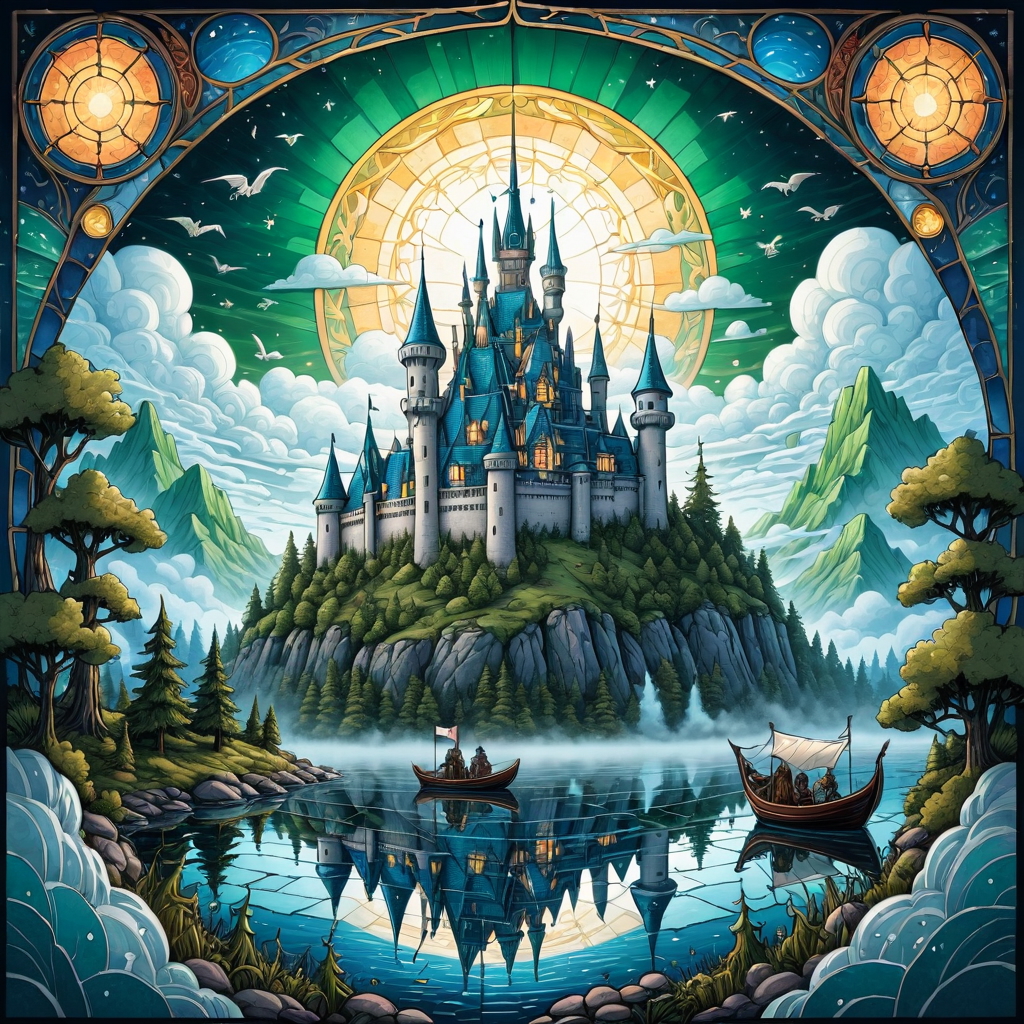}
    \end{subfigure}
    \hfill
    \begin{subfigure}[b]{0.32\linewidth}
        \centering
        \includegraphics[width=\linewidth]{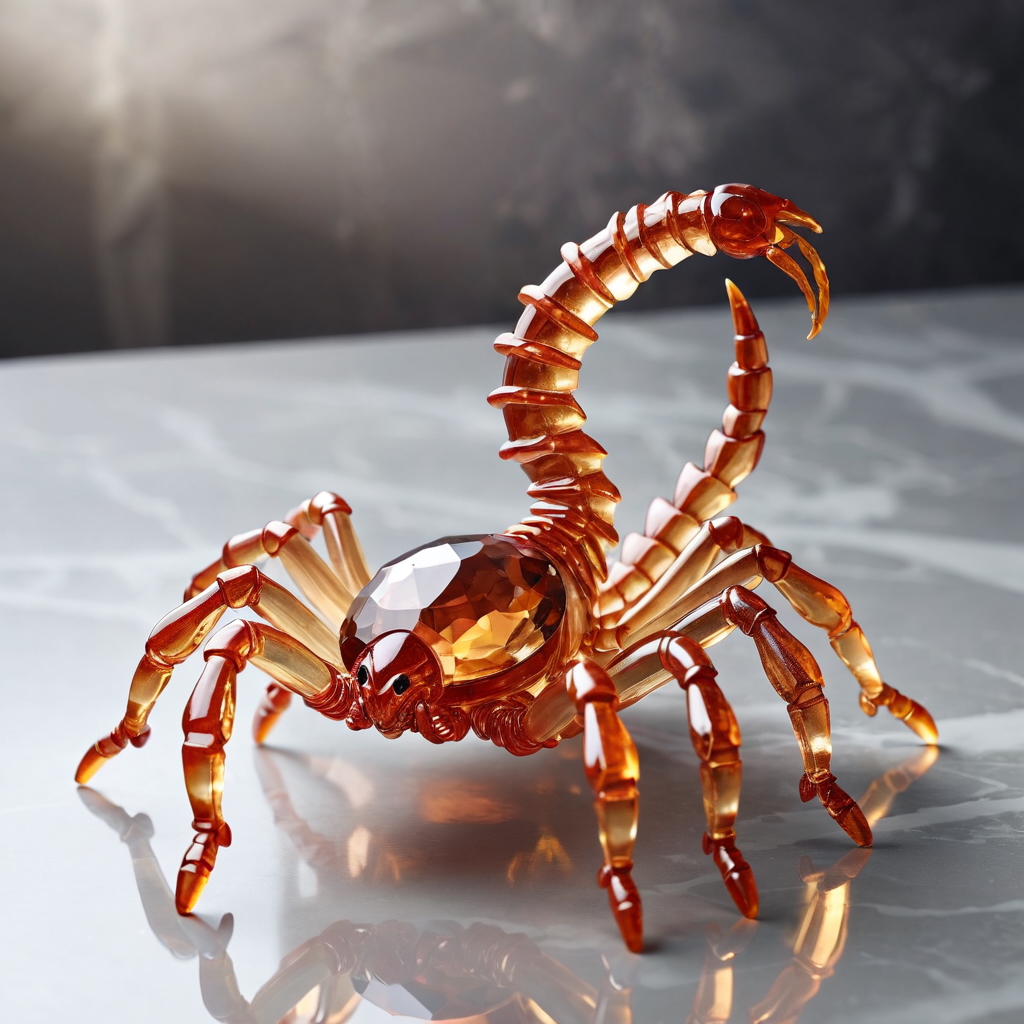}
    \end{subfigure}
    \hfill
    \begin{subfigure}[b]{0.32\linewidth}
        \centering
        \includegraphics[width=\linewidth]{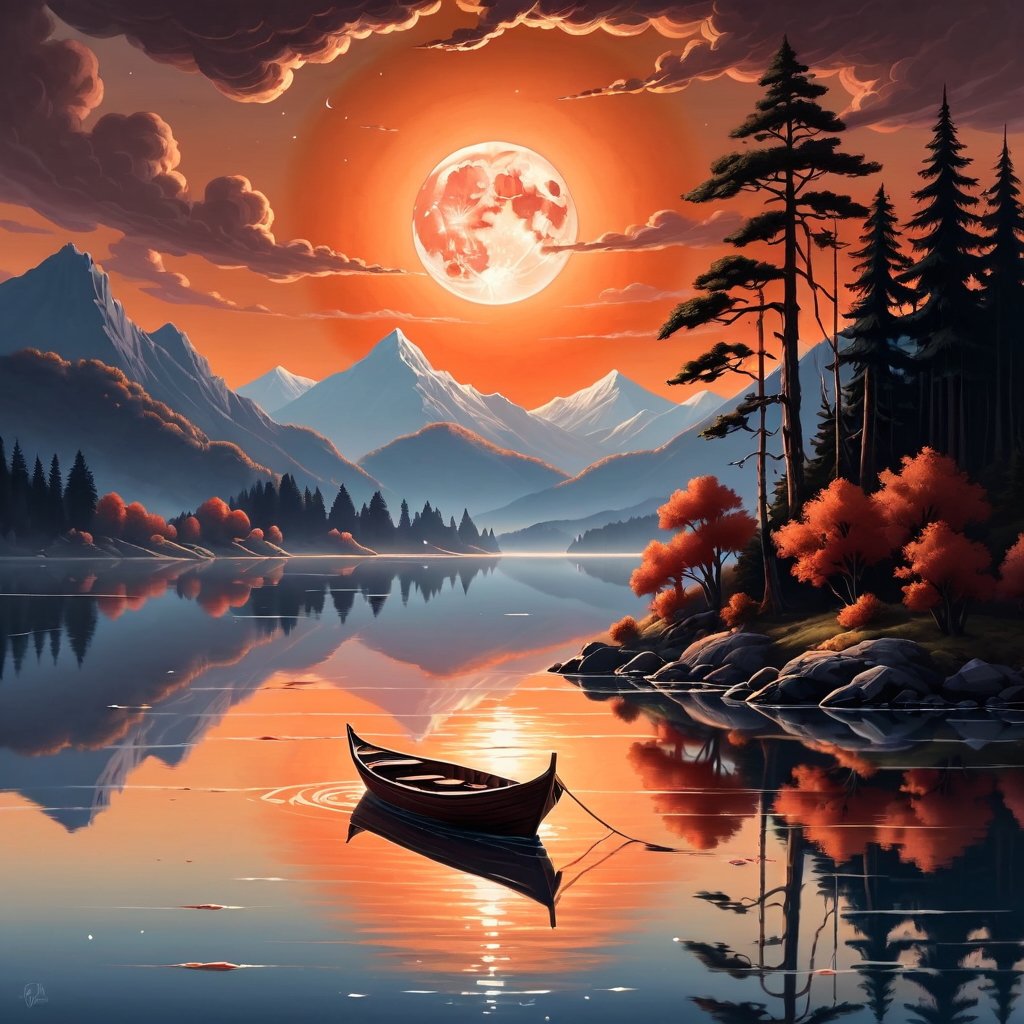}
    \end{subfigure}
\caption{More qualitative generations using \ourmethod{}}
    \label{fig:generations_2}
\end{figure}

\begin{figure}[ht]
    \centering
    \begin{subfigure}[b]{0.32\linewidth}
        \centering
        \includegraphics[width=\linewidth]{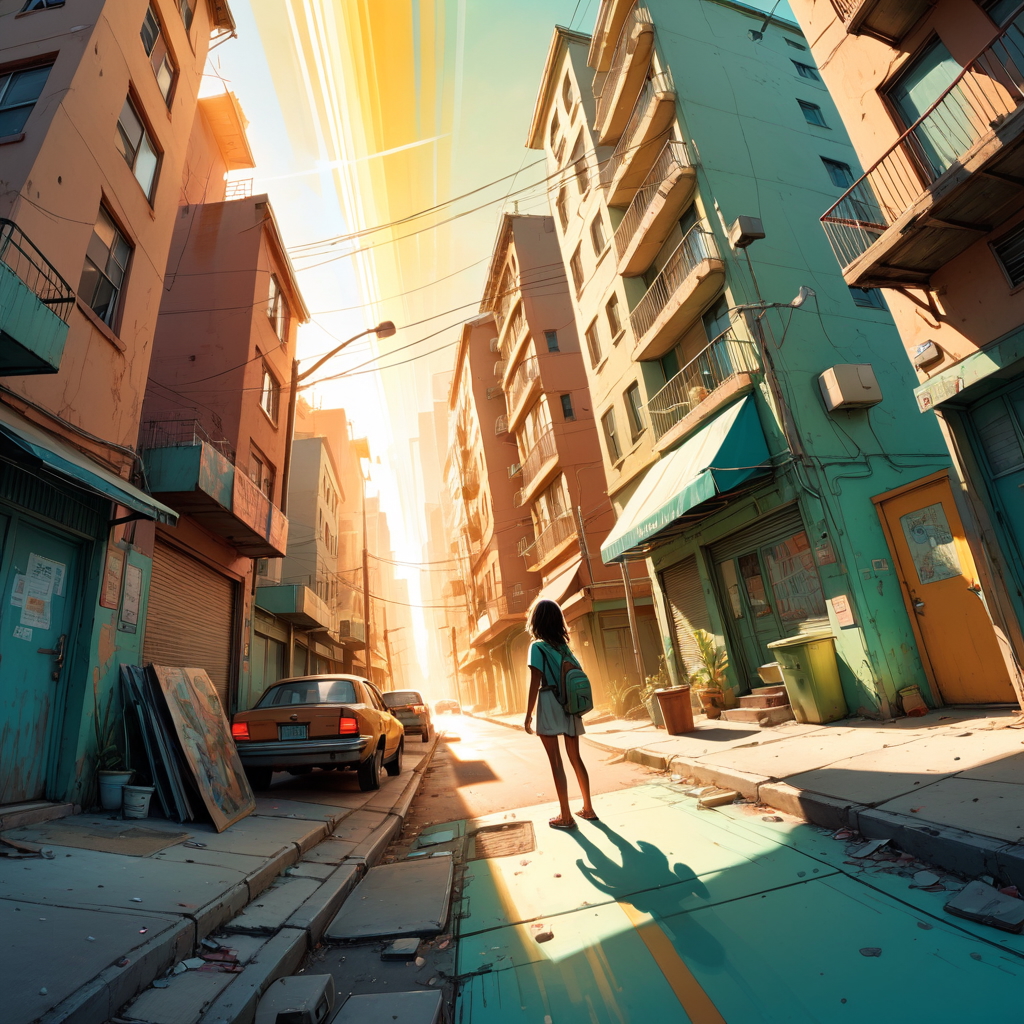}
    \end{subfigure}
    \hfill
    \begin{subfigure}[b]{0.32\linewidth}
        \centering
        \includegraphics[width=\linewidth]{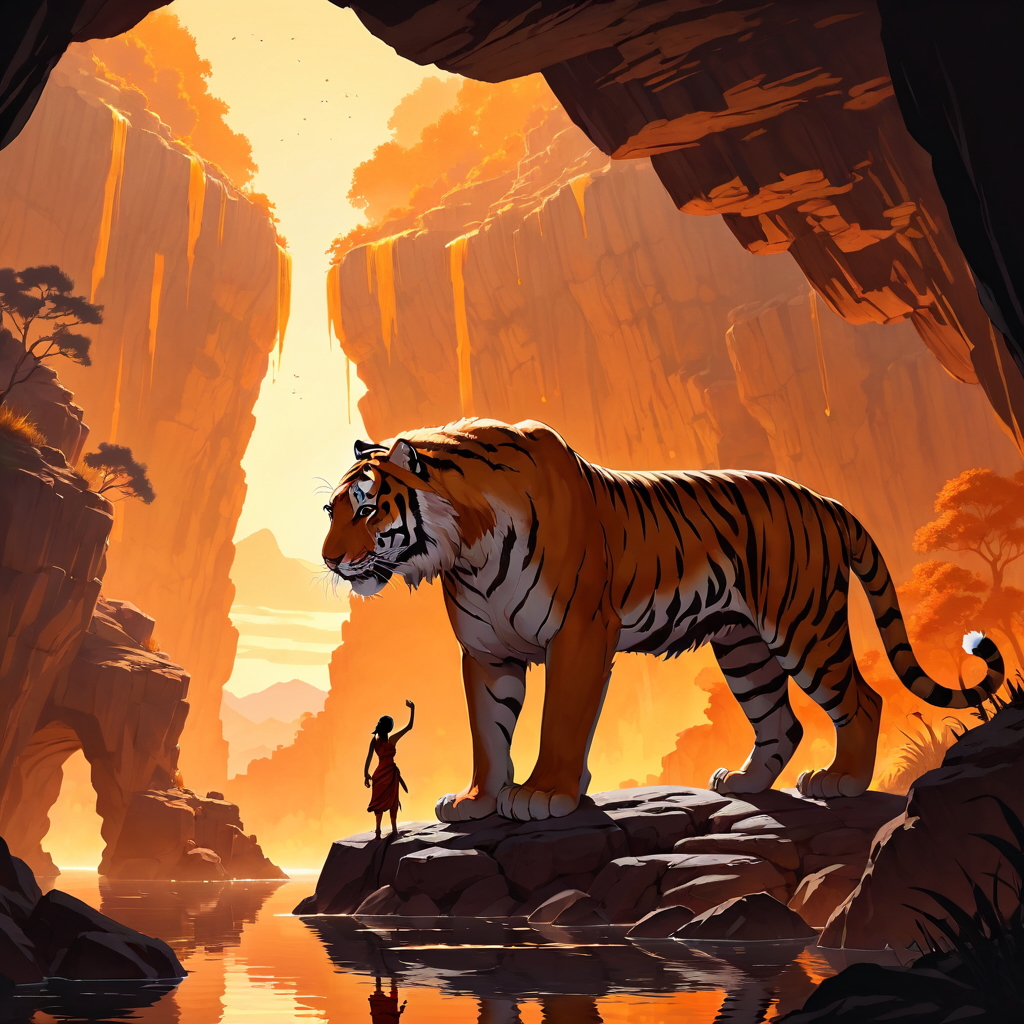}
    \end{subfigure}
    \hfill
    \begin{subfigure}[b]{0.32\linewidth}
        \centering
        \includegraphics[width=\linewidth]{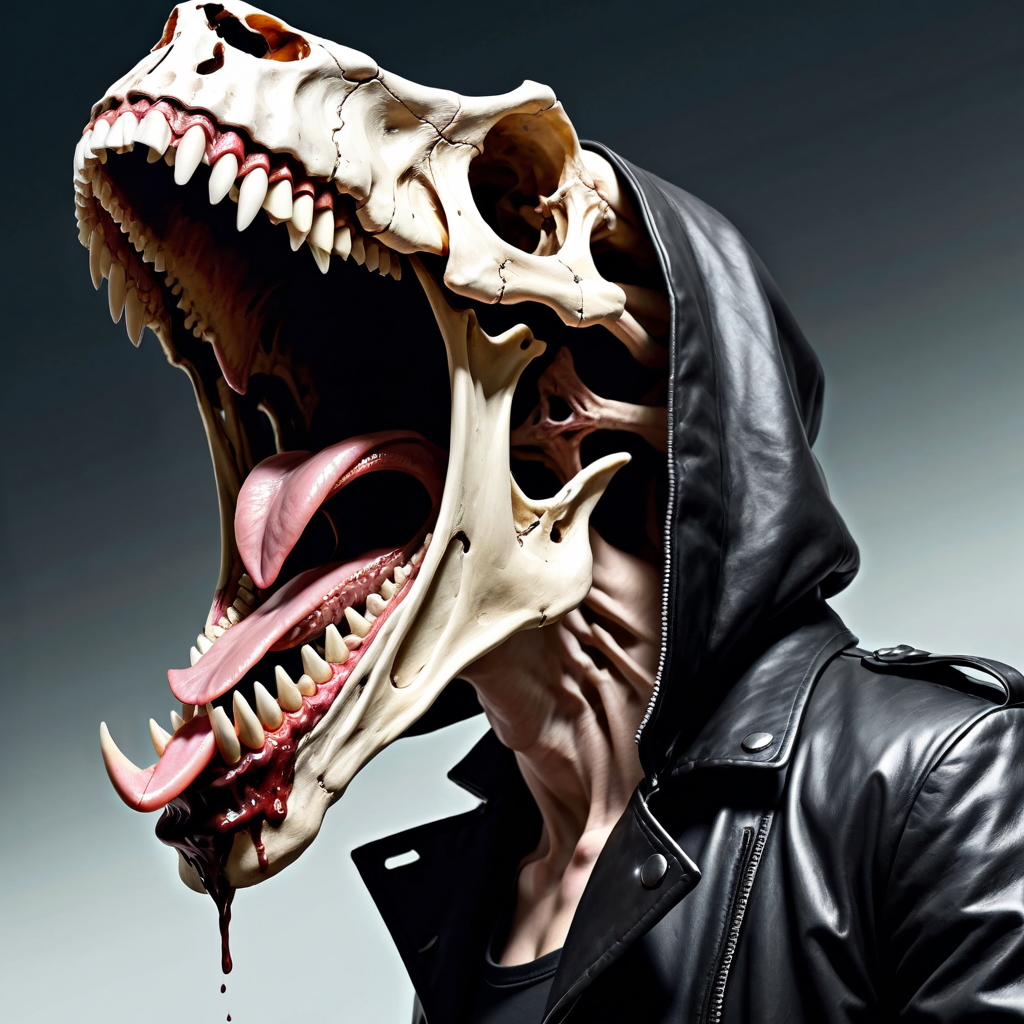}
    \end{subfigure}

    \vskip\baselineskip
    \begin{subfigure}[b]{0.32\linewidth}
        \centering
        \includegraphics[width=\linewidth]{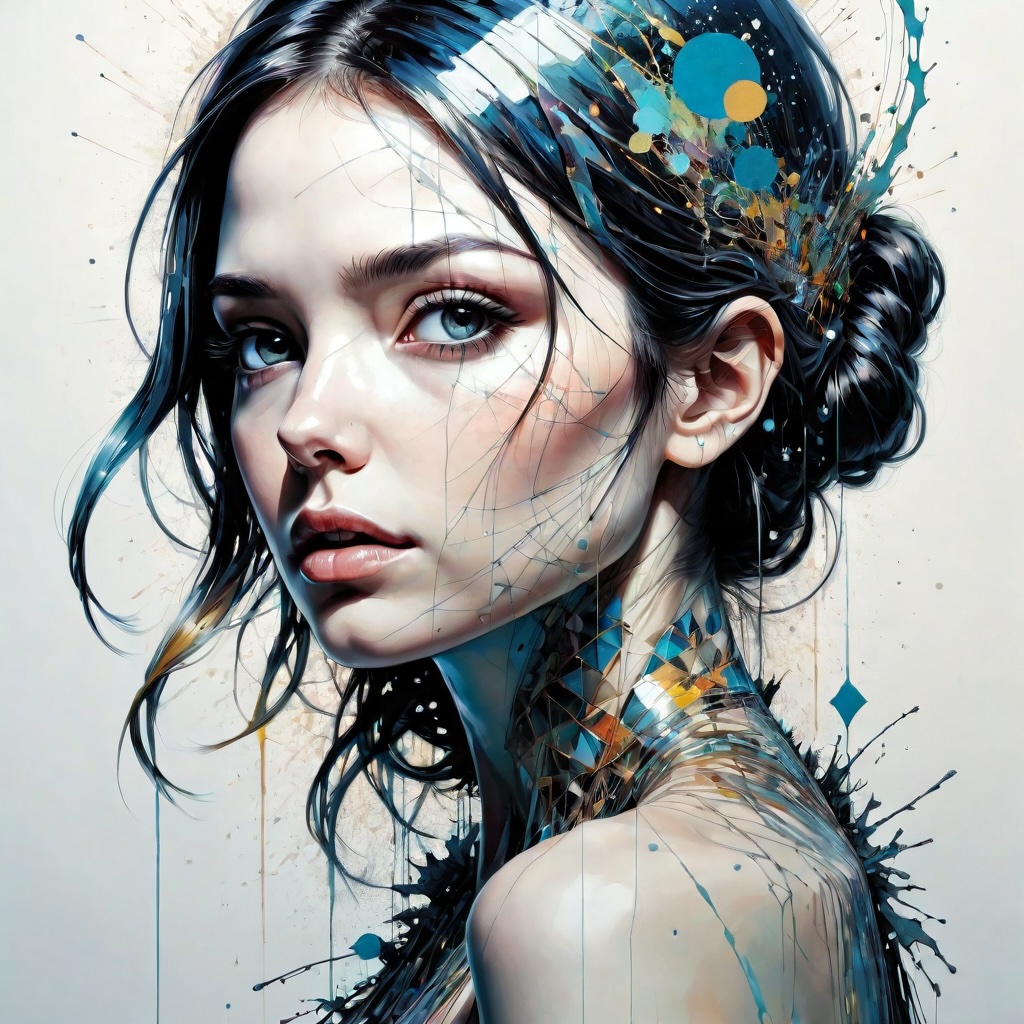}
    \end{subfigure}
    \hfill
    \begin{subfigure}[b]{0.32\linewidth}
        \centering
        \includegraphics[width=\linewidth]{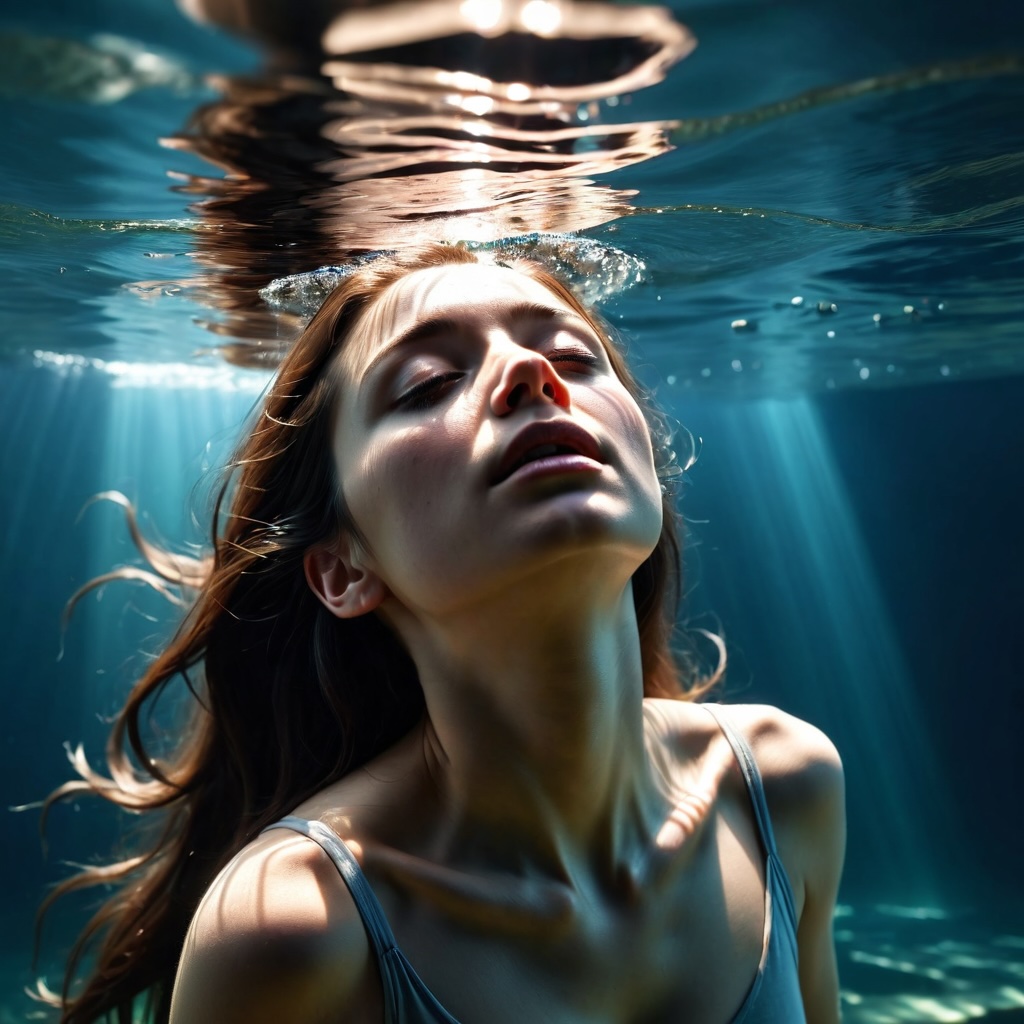}
    \end{subfigure}
    \hfill
    \begin{subfigure}[b]{0.32\linewidth}
        \centering
        \includegraphics[width=\linewidth]{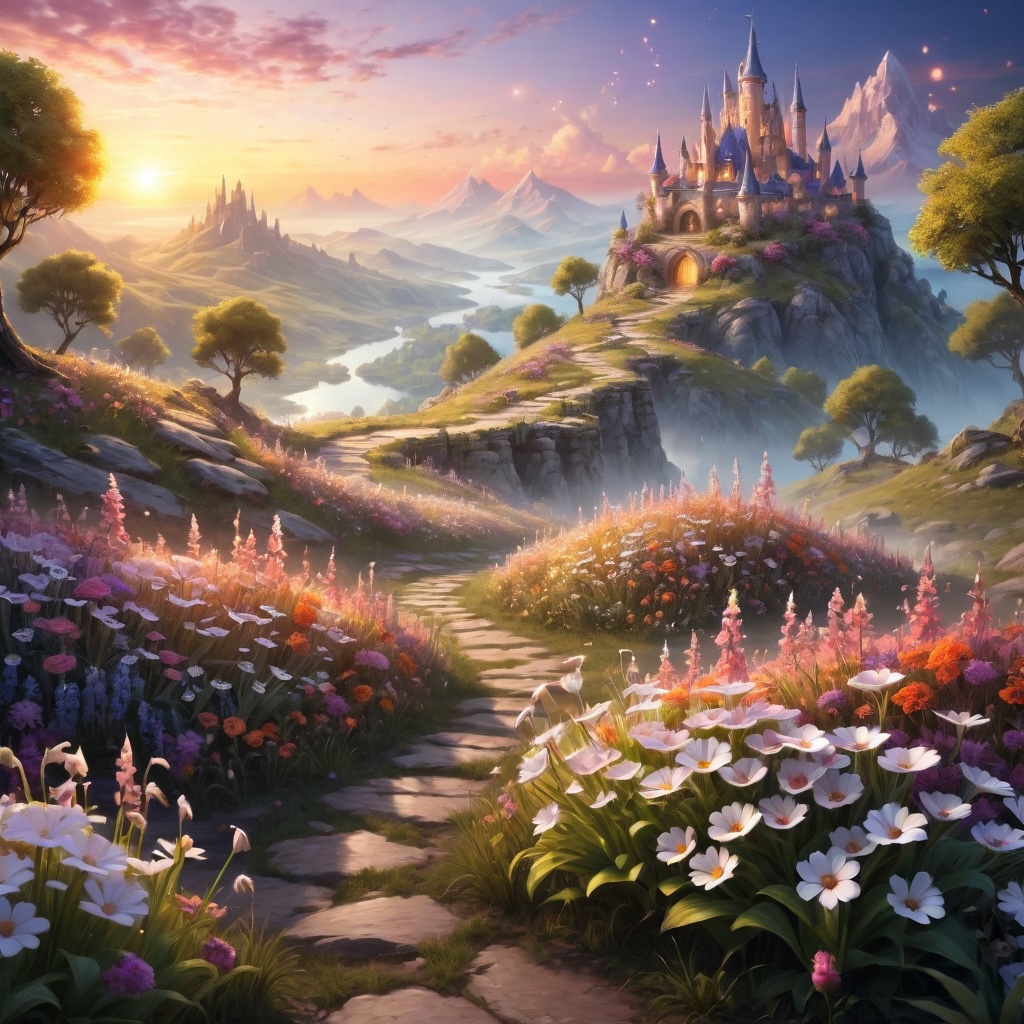}
    \end{subfigure}

    \vskip\baselineskip
    \begin{subfigure}[b]{0.32\linewidth}
        \centering
        \includegraphics[width=\linewidth]{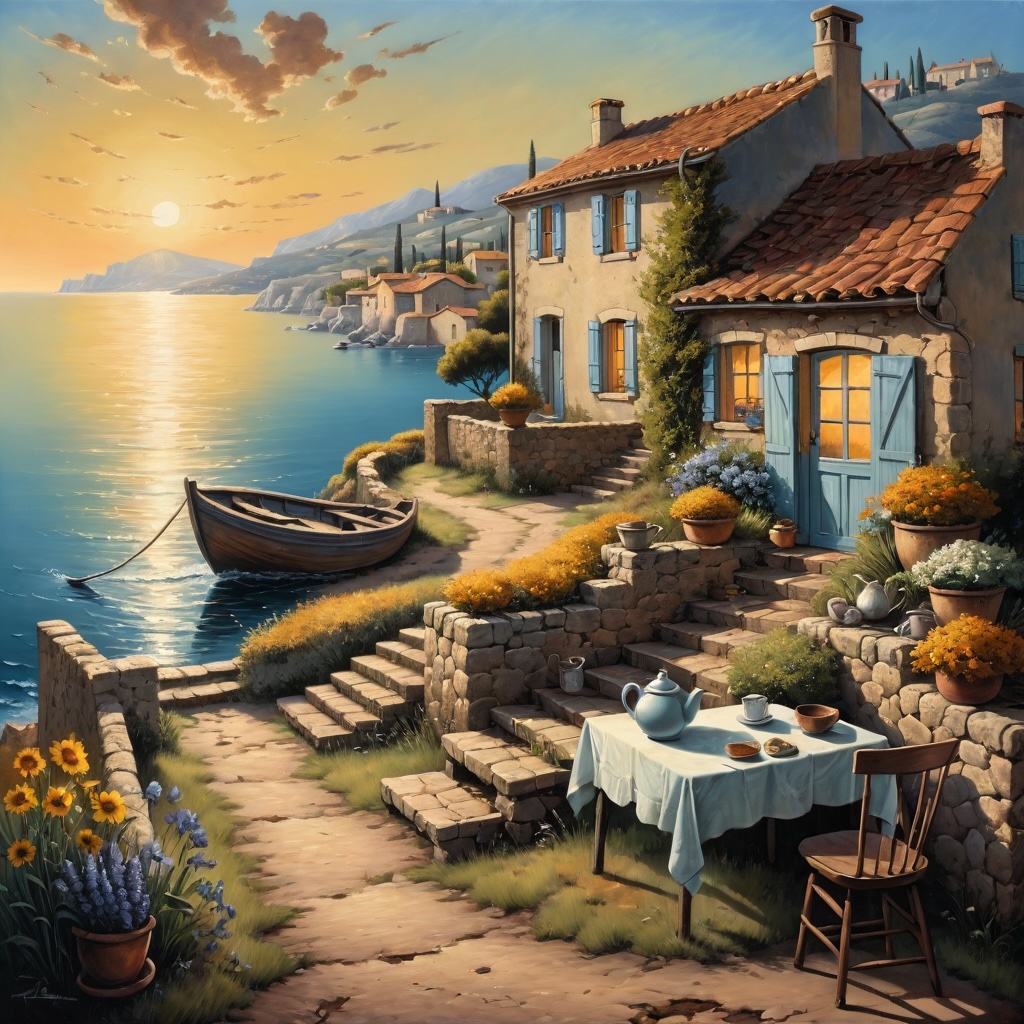}
    \end{subfigure}
    \hfill
    \begin{subfigure}[b]{0.32\linewidth}
        \centering
        \includegraphics[width=\linewidth]{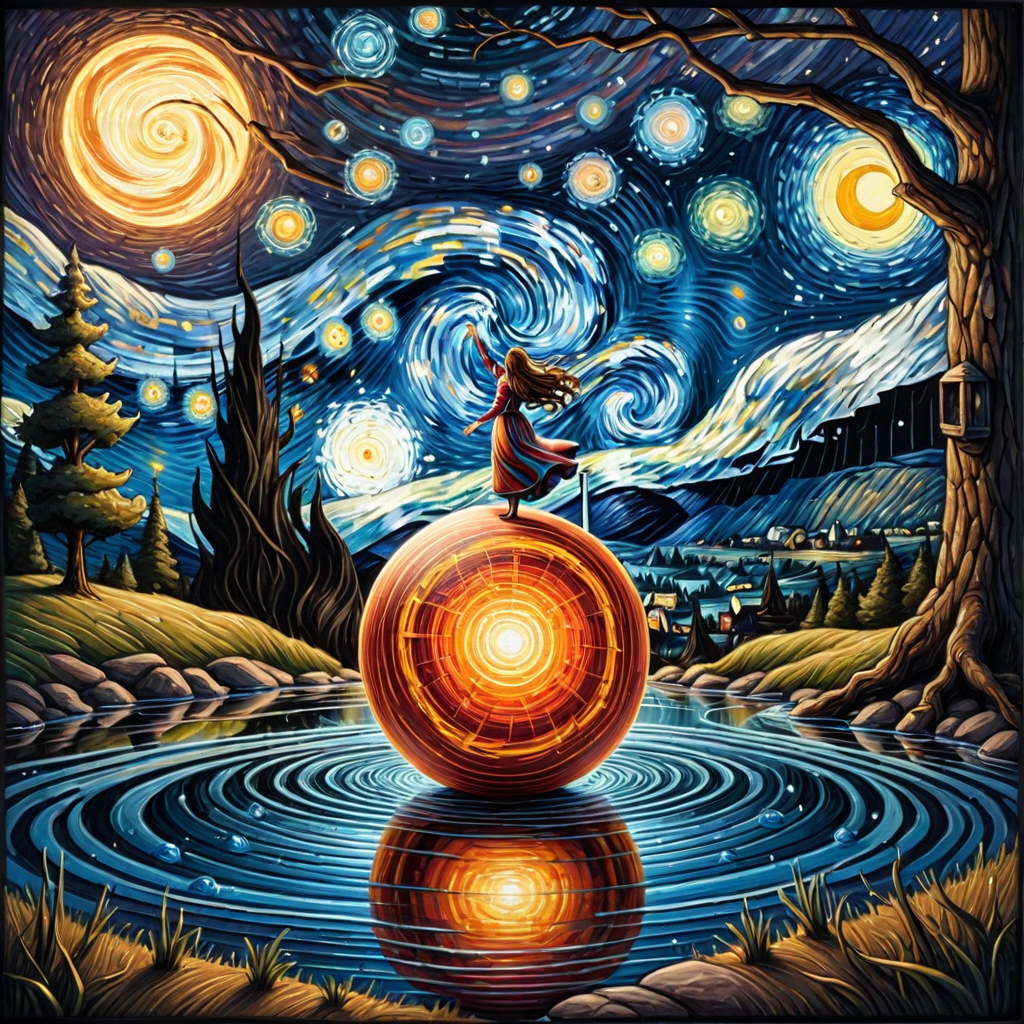}
    \end{subfigure}
    \hfill
    \begin{subfigure}[b]{0.32\linewidth}
        \centering
        \includegraphics[width=\linewidth]{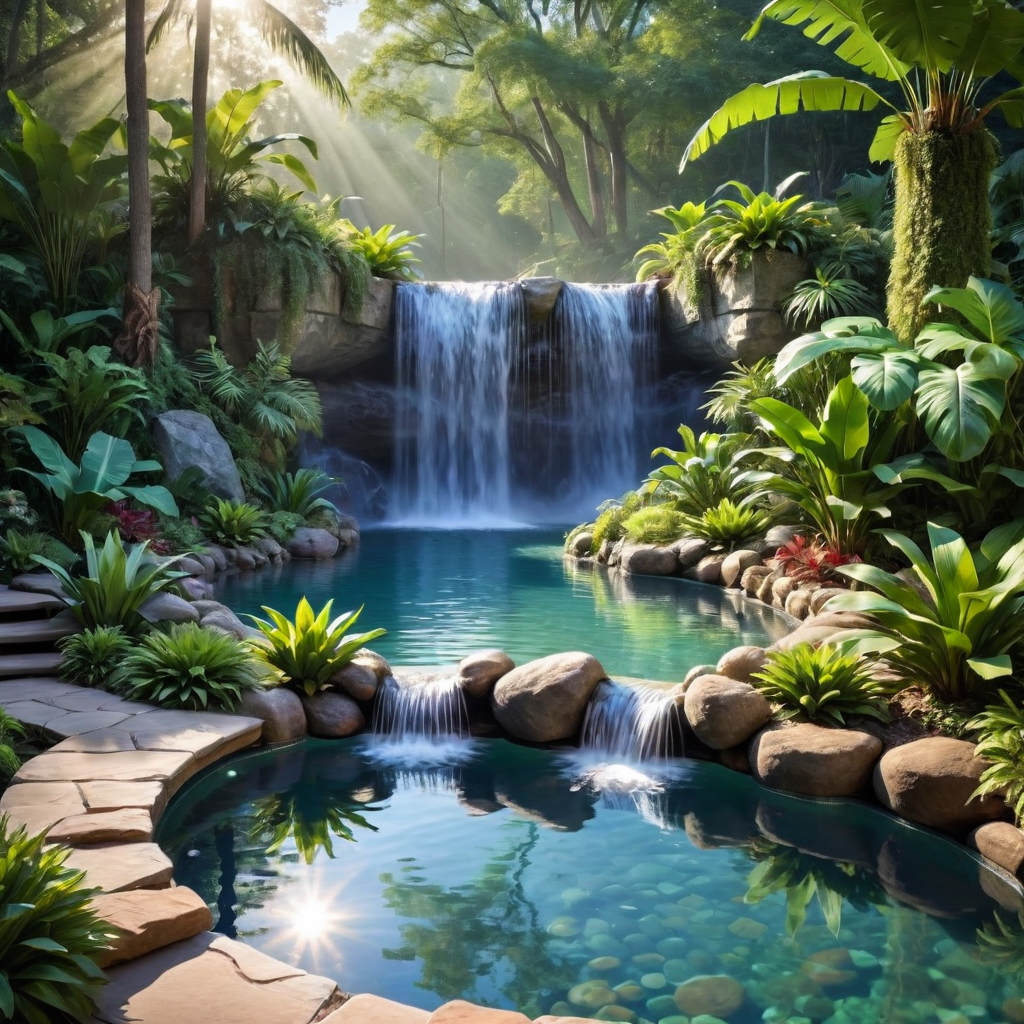}
    \end{subfigure}

    \vskip\baselineskip
    \begin{subfigure}[b]{0.32\linewidth}
        \centering
        \includegraphics[width=\linewidth]{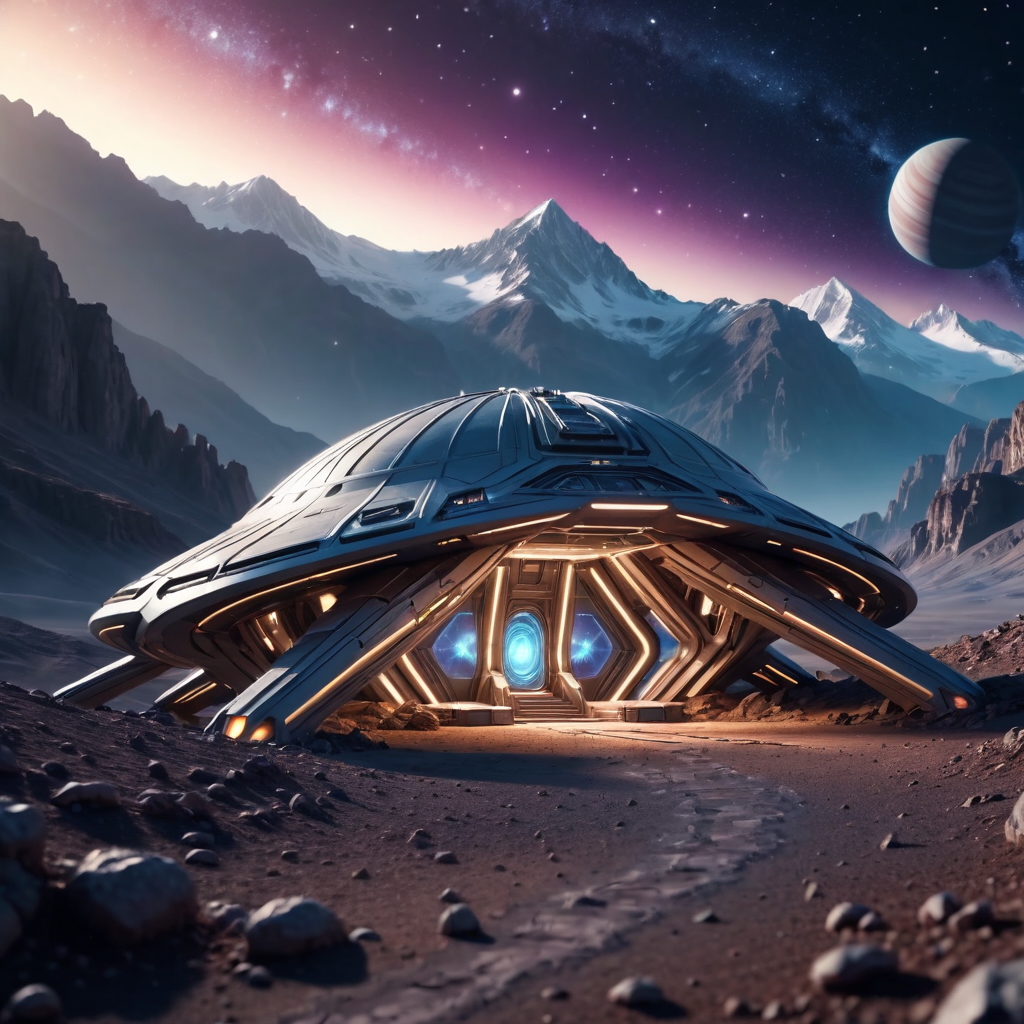}
    \end{subfigure}
    \hfill
    \begin{subfigure}[b]{0.32\linewidth}
        \centering
        \includegraphics[width=\linewidth]{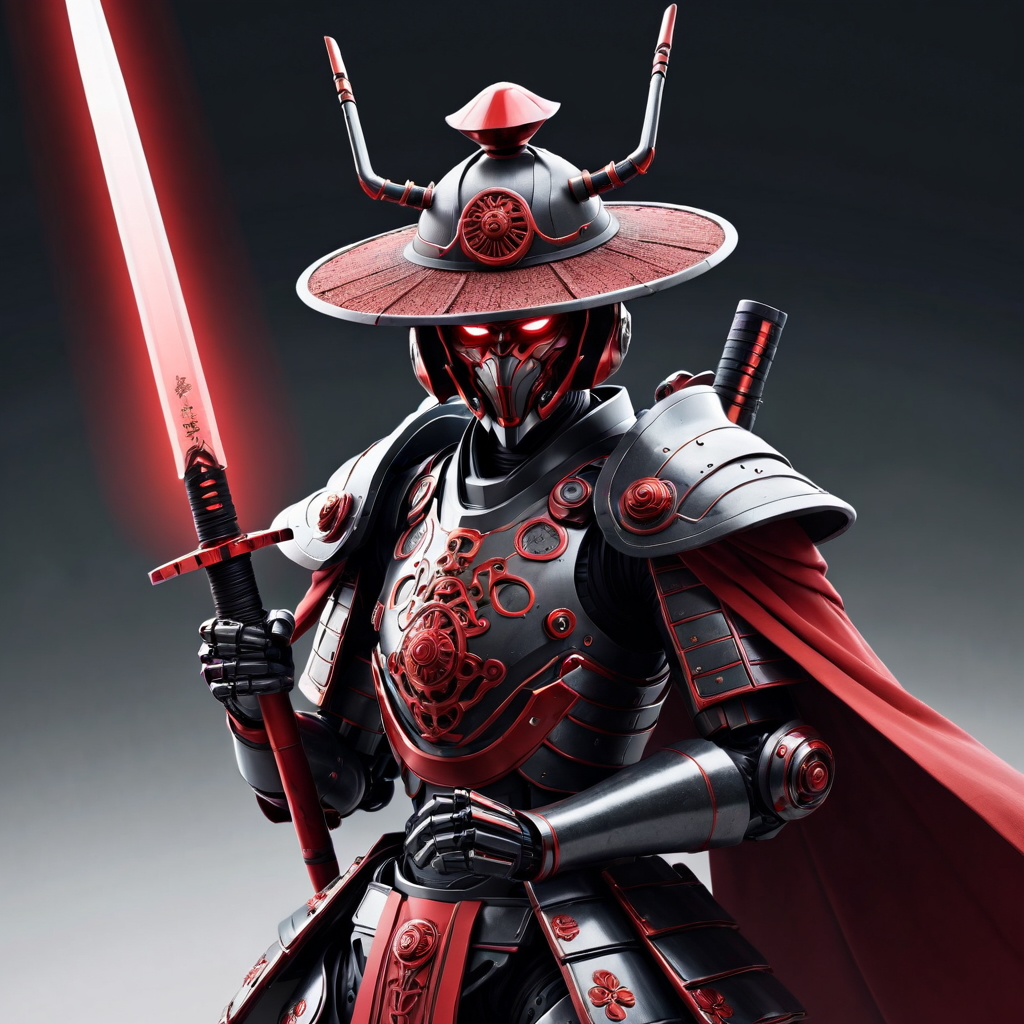}
    \end{subfigure}
    \hfill
    \begin{subfigure}[b]{0.32\linewidth}
        \centering
        \includegraphics[width=\linewidth]{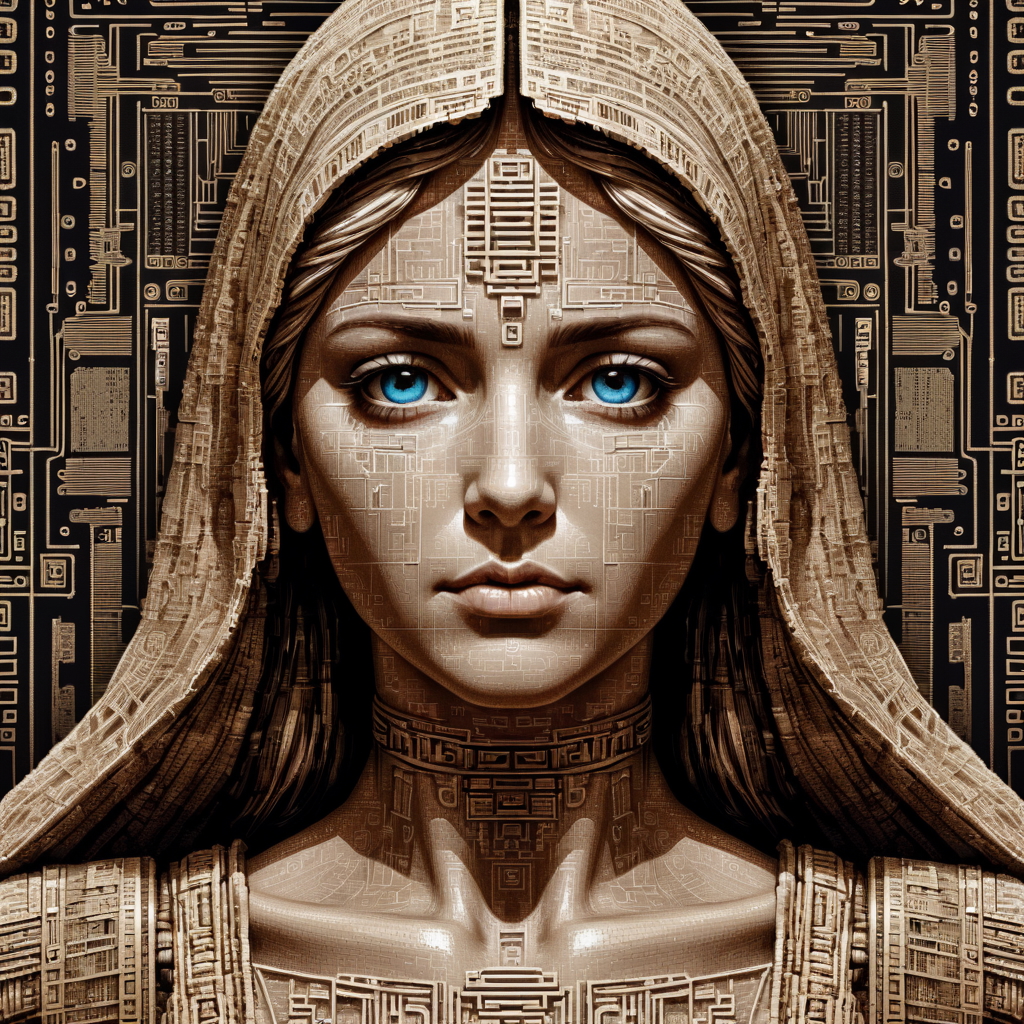}
    \end{subfigure}
    \caption{More qualitative generations using \ourmethod{}}
    \label{fig:generations_1}
\end{figure}

Figures ~\ref{fig:generations_3},~\ref{fig:generations_2} and~\ref{fig:generations_1} provide additional generations of CivitAI prompts using \ourmethod{}. We give a detailed list of the relevant prompt (ordered by appearance order, from top-left to bottom right)

\begin{figure}[!htb]
    \centering
    \includegraphics[width=1\linewidth]{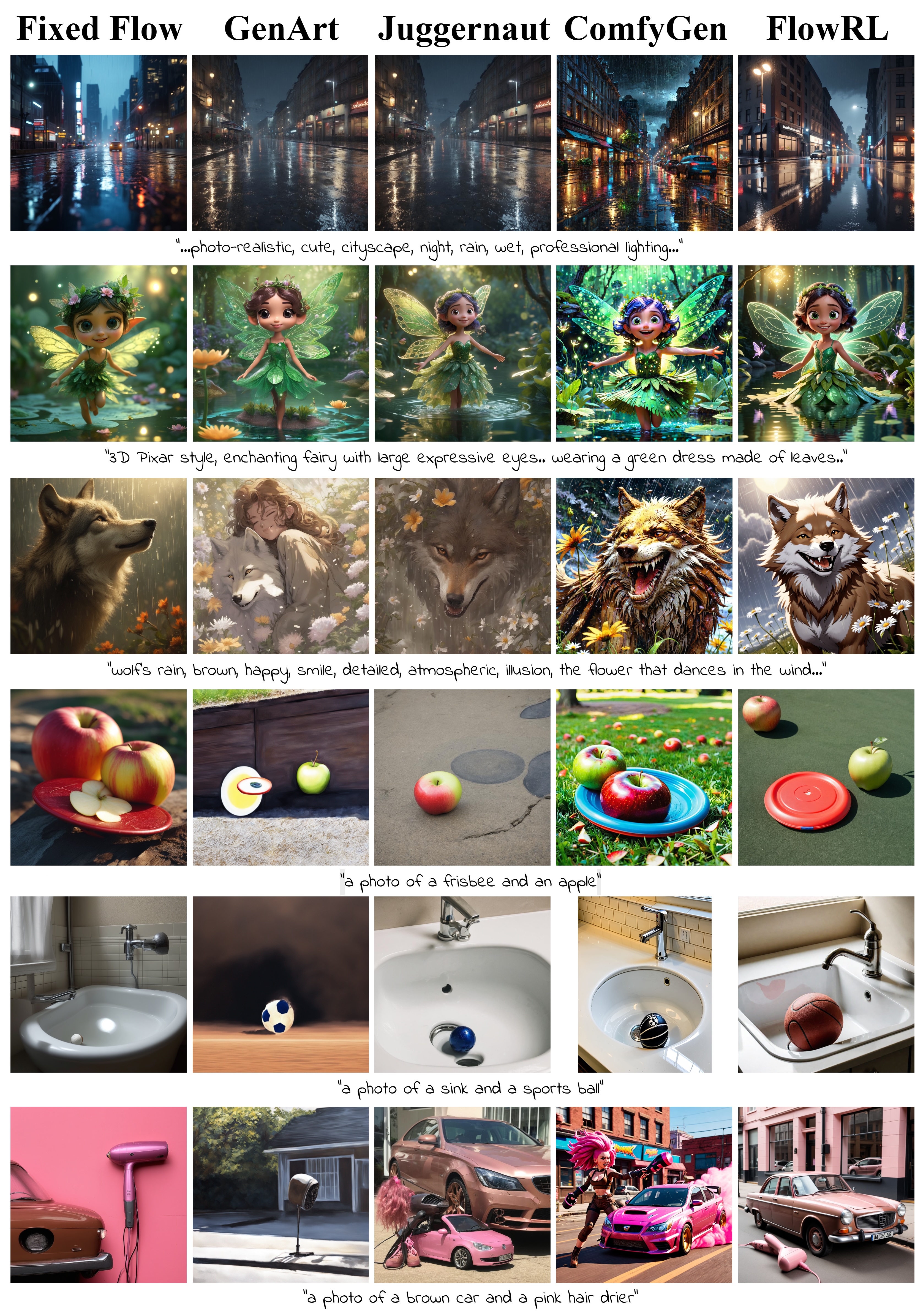}
    \caption{Additional qualitative comparisons on CivitAI prompts (top 3) and GenEval prompts (bottom 3)}
    \label{fig:more_civit}
\end{figure}

We provide additional qualitative comparisons between \ourmethod{} and the baselines in Figure~\ref{fig:more_civit} for both CivitAI prompts and GenEval prompts.

\begin{figure}[ht]
    \centering
    \includegraphics[width=1\linewidth]{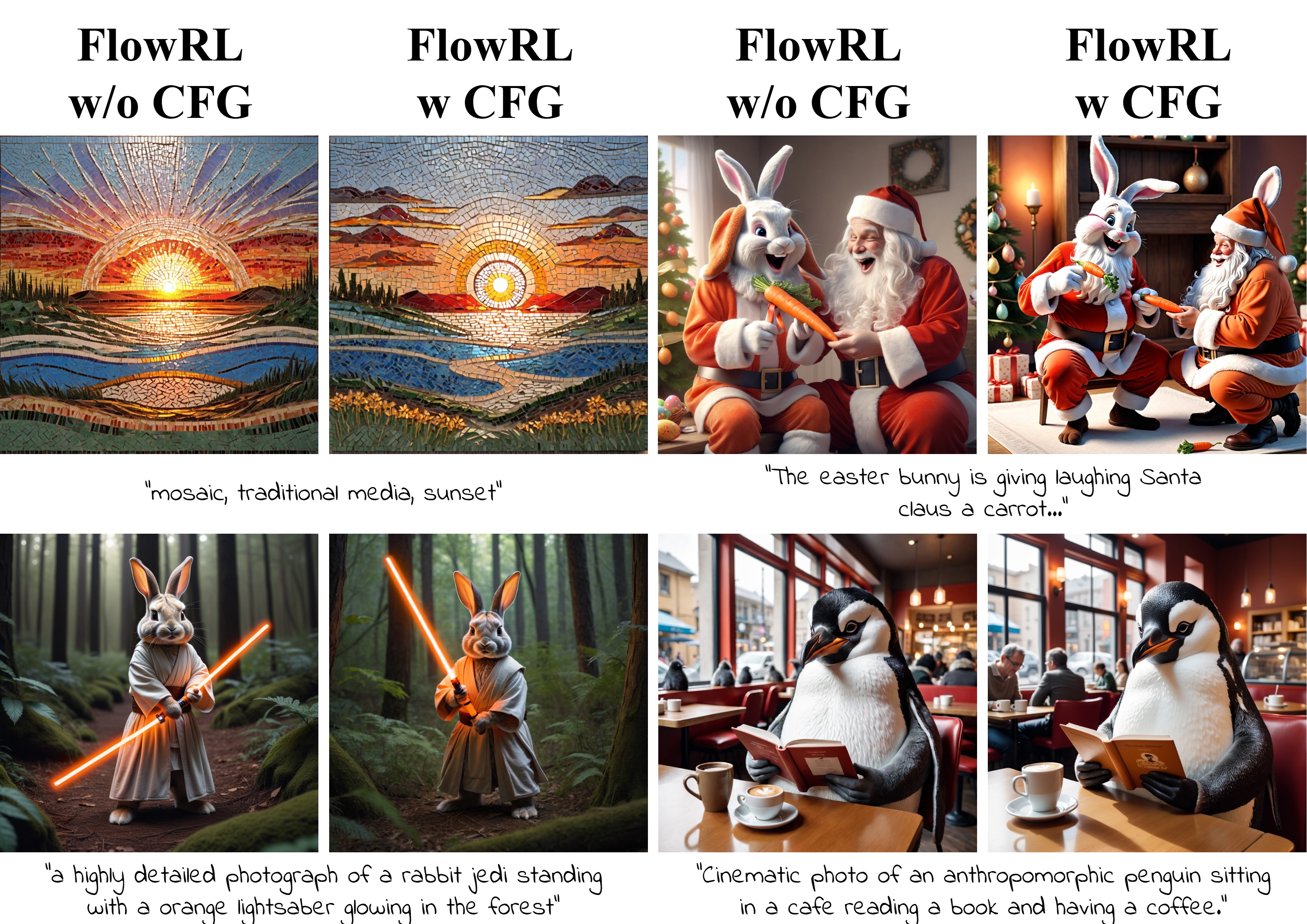}
    \caption{Qualitative example on influence of CFG on the output image}
    \label{fig:cfg}
\end{figure}

In addition in figure ~\ref{fig:cfg} we give a qualitative comparison between \ourmethod{} with and w/o the usage of the dual model guidance mechanism (CFG).

\paragraph{List of prompts for example generations}
\begin{enumerate}
    \item "Amazing detailed photography of a cute adorable samurai kitten holding Katana with 2 paws, Cherry Blossom Tree petals floating in air, high resolution, piercing eyes, lifelike fur, Anti-Aliasing, FXAA, De-Noise, Post-Production, SFX, insanely detailed \& intricate, hypermaximalist, elegant, ornate, hyper realistic, super detailed, noir coloration, serene, 16k resolution, full body"
    \item "masterpiece, best quality, high quality, intricate, absurdres, very aesthetic, no humans, landscape, outdoors, mountain tops, wind, windy, wind lines, clouds, above clouds, cliff, wind magic, aurora, ultra wide angle shot, cinematic style, highly detailed, extremely detailed, sharp detail, majestic, shallow depth of field, movie still, soft light, circular polarizer, colorful, wallpaper, professional illustration, anime"
    \item "pixar style of turtle, as a pixar character, tinny cute, luminous, wearing hawaiian hat, at the sea shore, tropical beach, smile, high detailed, photorealistic, 8k"
    \item "Medieval German castle, surrounded by mountains, high fantasy, epic, digital art."
    \item "style of Edvard Munch, Piercing, sagacious eyes, mirage-like, the Sandswept dreamdweller, a trickster of dunes, clad in a wind-whispered turban, eternally smirking, sandswaggling over a dune-freckled miragepath in an ancient zephyr-twisted cactidle wilderness of towering dustfrond phantasmagorias, paying no heed to the sun-scorched skyripples above, Arid, Sand-whirled, Mirage, Cacti, Mystical Desert, oasis illusions. Edvard Munch style"
    \item "full body, Fat cats at Elrond's council from the movie Lord of the Rings, fluffy paws, background action-packed"
    \item "detailed, vector art, thick lines, oil painting, vibrant, colorful, candy pink, scarlet red, orange, smooth coloring, nature, landscape, stone pillars, long wild trees, moody streaks sky, natural lighting, river, reflections, best composition, background"
    \item a woman with red hair and a white shirt is shown in this painting style photo with a pink background, Charlie Bowater, stanley artgerm lau, a painting, fantasy art masterpiece, best quality, depth of field, backlighting, intricate details"
    \item "cinematic shot of stone giant walking in lush forest, dappled sunlight, high resolution"
    \item "Majestic jagged rocky mountains, red mesas, wind eroded colorful rock formations, twighlight, starry night, petrified forest national park, arizona, astrophotography"
    \item "Cubist inspiration, A landscape represented with planes and flat colors. The landscape could show a field, forest or city, and flat planes and colors could be used to create a sense of depth and perspective, surrealism, aesthetic, bold gorgeous colours, high definition, super clear resolution, iridescent watercolor ink, acid influence, fantastic view, crisp quality, complex background, medium: old film grain, tetradic colors, golden hour, rust style, vantablack aura, golden ratio, rule of thirds, cinematic lighting Dark realism and magical. Complementary poisonous colors with deep zoom Memphis style abstract bokeh background with deep zoom"
    \item "FrostedStyle Highly detailed Dynamic shot of a transparent frosted ruby reindeer, glowing with rage from within extremely detailed"
    \item "vertical symmetry, vntblk, movie poster art, blood moon, red moon, darkest night, stonehenge, low angle:famous artwork by caspar david friedrich and stephan martiniere, perfectly round scifi portal, ominous dark surreal and unique landscape with towering obelisks piercing the sky, glowing ornate lovecraftian artifact, jagged rock formations, night sky, mysterious, ethereal, deserted, dark corners, burgundy, anthrazit grey, crimson, sunset orange, yellow, teal:16, ultra detailed"
    \item "by Peter Holme III and Roger Dean and Vitaly Golovatyuk and Mark Lovett, cinematic, shallow depth of field"
    \item "grainy, extremely detailed, intricate detail, dynamic lighting, photorealistic, filmg, natural lighting, low light, cat, slime, red glowing eyes, :P, fluffy, hairy, fluff, glowing stripes, raining, wet, dark theme, open mouth, lot of teeth, abyss, lurking in shadow"
    \item "The art of Origami, Paper folding, Swan on a lake, Amazing colours, Intricate details, Painstaking Attention to Details, UHD"
    \item "amateur analog photo, The creature monster brown fur Easter bunny character covered in yeast, evil, creepy, in dark forest, fine textures, high quality textures of materials, volumetric textures, natural textures"
    \item "In a wondrously gleaming futuristic realm composed entirely of ripe peaches, a towering palace made of glistening peach flesh and pitted stone stands as the focal point of the image. The palace's walls are adorned with intricate carvings of peach vines and blossoms, while peach juice flows like streams through the city streets. This vivid and surreal painting captures the ethereal beauty of a world where nature and architecture are seamlessly intertwined, every detail rendered with unparalleled precision and depth, making viewers feel as if they could reach out and touch the succulent fruit structures."
    \item "high-contrast palette, cinematic quality, fashion photography, chimp wearing a black suit with a black shirt with a black vest with a black necktie with black Rayban style sunglasses, natural skin texture, realistic skin texture, skin pores, skin oils"
    \item "faistyle, retro artstyle, painting medium, lake, mountain, forest"
    \item "close up Portrait photo of muscular bearded guy in a worn mech suit, light bokeh, intricate, steel metal rust, elegant, sharp focus, photo by greg rutkowski, soft lighting, vibrant colors, masterpiece, streets, detailed face"
    \item "detailed ink, pen and ink, mail art, best quality, detailed epic ice transparent ethereal otherworldly ghost castle in the blue sky, clouds, smoke, fog, detailed landscape, ghost figures, lake, boat, green forest, detailed flying dragon at the sky, detailed scales, warm lights, glittering, Craola, Dan Mumford, Andy Kehoe, 2d, flat, art on a cracked paper, patchwork, stained glass, cute, adorable, fairytale, storybook detailed illustration, cinematic, ultra highly detailed, tiny details, beautiful details, mystical, luminism, vibrant colors, complex background"
    \item "crystal scorpion"
    \item "the image portrays a tranquil scene of a boat floating gently on the water, surrounded by an expansive landscape. the moon, full and glowing with a warm, reddish orange hue, casts a mystical ambiance over the entire scene. its reflection shimmers off the surface of the water, adding to the serene atmosphere. in the distance, mountains loom under the moon's soft glow, their peaks partially obscured by the low hanging clouds. they appear majestic yet gentle, as if watching over the peaceful night below. trees line the shore in the foreground, their silhouettes faintly visible against the darkening sky. this picturesque setting evokes a sense of calm and tranquility, inviting viewers to take a moment and appreciate the beauty of nature. it is a symphony of colors and shapes, each element working harmoniously together to create a visually captivating and emotionally soothing composition."
    \item "hyper detailed, elusive, exotic, angelic, luminescent, by James Gilleard and by Alice Pasquini, point-of-view shot, fisheyes view, sunbeams lighting"
    \item "A colossal majestic tiger, looms over a cavern's silhouette, gazing intently at a small human figure with a stance of curiosity, the figure a silhouette against a backdrop bathed in the warm oranges and yellows of a sun, faces the tiger unafraid, with floating embers dancing around them both, scene of serene confrontation amidst the enveloping dusk"
    \item "illustration, solo, animal skull head, screaming, long split tongue, fangs, head closeup, black leather coat, horror atmosphere, side view"
    \item "mysterious silhouette of woman from the enchanted pond, abstract art, by Minjae Lee, Carne Griffiths, Emily Kell, Geoffroy Thoorens, Aaron Horkey, Jordan Grimmer, Greg Rutkowski, extraordinary depth, masterpiece, surreal, geometric patterns, extremely detailed, bokeh, perfect balance, deep and thin edges, artistic photorealism, smoothness, excellent masterpiece by the head of rapid engineering, white background: 1.2"
    \item ""Struggling to breathe, like being held under water" beautiful inner light, deep shadows, extraordinary detail"
    \item "a fantasy landscape at dawn covered in magical flowers"
    \item "vintage, shabby, morning, dawn, cozy world, Kruskamp, Monge, Kincaid, Potter, Dali, Burton, oil, coal, provence, house by the sea, cozy and beautiful landscape, double composition, drama, tragedy, the core of magic"
    \item "fine art, oil painting, best quality, dark tales, illustration, each color adds depth, and the entire piece comes together to create a breathtaking spectacle of motion and tranquility., while the ball is adorned with an array of stripes in various hues. the figurine, while her right hand delicately holds a small, epic splash cover art in the van gogh style, starry sky, dan mumford, andy kehoe, 2d, flat, delightful, vintage, art on a cracked paper, patchwork, stained glass, fairytale, storybook detailed illustration, cinematic, ultra highly detailed, tiny details, beautiful details, mystical, luminism, vibrant colors, complex background"
    \item "Envision a breathtaking waterfall cascading into a crystal-clear pool surrounded by lush greenery. The pool is home to magical water creatures, including playful water sprites and elegant swans with feathers that shimmer in shades of silver and gold. Mist rises from the waterfall, creating rainbows in the sunlight. Curious frogs with iridescent skin leap from rock to rock, while dragonflies with jeweled wings flit above the water. The sanctuary is a hidden paradise, inviting all who enter to experience the tranquility and magic of this enchanting world"
    \item "futuristic building, surface from a alien planet, mountains in the background, sci fi, fantasy, space art, galaxy background, shotting stars, dynamic angle, intricate "
    \item "Image is a digital artwork featuring a futuristic samurai robot. The robot has a sleek, metallic body with intricate mechanical details and a predominantly black and silver color scheme. It wears a large, red, conical hat and a matching red cape that flows behind it. The robot's face is obscured by a mask, giving it a mysterious appearance. It holds a red and black katana in its right hand, ready for combat. The background is a gradient of dark grey, with a circular, smoky effect behind the robot, adding to the dramatic and intense atmosphere of the scene."
    \item "masterpiece, ASCII, 8k.absurdes, intricate, maximum resolution, hyper detailed, Mirage, DonMn1ghtm4reXL, glow, fog, obsidian armor with red ruby, details, hellgate london themed, demoniac armor, force huge demoniac wide wings, glowing wings, energy wings"
\end{enumerate}

\clearpage

\subsection{Tokenization Encoding Method}
We developed a systematic procedure to transform JSON-based workflow representations into a compact, encoded format. This process utilizes schema learning to ensure both accuracy and efficiency in data transformation.

\paragraph{Methodology}
First, we infer a schema from a collection of workflow JSON files by iterating through each file and extracting the class types and field information for all utilized nodes. The resulting schema is stored for future encoding tasks. In the encoded representation, nodes are sorted and formatted to include their class types and input values, with explicit references to connected nodes. Each line in the encoded output corresponds to a node from the original JSON structure, providing a clear and organized mapping.

\paragraph{Incorporating Workflow-Specific Tokens}
To more effectively capture the structure of ComfyUI workflows, we enhanced the base tokenizer by introducing custom tokens that represent key workflow elements such as node types, connections, and parameters. This enriched tokenization scheme helps the model better understand relationships between workflow components.
Below, we provide examples of some of the custom tokens added to the tokenizer:
\begin{itemize}
\item ng Everywhere3
\item AspectSize
\item Automatic CFG
\item BNK\_AddCLIPSDXLRParams
\item BNK\_CLIPTextEncodeAdvanced
\item BasicPipeToDetailerPipe
\item Image Levels Adjustment
\item Image Remove Background (rembg)
\item CLIP Positive-Negative XL w/Text (WLSH)
\item CLIP=
\item CLIPLoader
\item CLIPMergeSimple
\item CLIPSetLastLayer
\item CLIPTextEncode
\item CLIPTextEncodeSDXL
\item CLIPTextEncodeSDXLRefiner
\item CLIP\_NEGATIVE
\item CONDITIONING=
\item CR Apply LoRA Stack
\item CR Apply Model Merge
\item SDXL 1.0/animagineXLV31\_v30.safetensors
\item SDXL 1.0/crystalClearXL\_ccxl.safetensors
\item SDXL 1.0/dreamshaperXL\_turboDpmppSDEKarras.safetensors
\item SDXL 1.0/envyhyperdrivexl\_v10.safetensors
\item SDXL 1.0/faces\_v1.safetensors
\item SDXL 1.0/jibMixRealisticXL\_v90BetterBodies.safetensors
\item SDXL 1.0/juggernautXL\_v9Rdphoto2Lightning.safetensors
\end{itemize}

\subsubsection{Example of encoded flow}
Following is a short example of the original ComfyUI flow in it's JSON form and it;s encoded text form:

\paragraph{JSON Form}
\begin{verbatim}
{
  "2": {
    "inputs": {
      "ckpt_name": "SDXL 1.0/realvisxlV40_v40LightningBakedvae.safetensors"
    },
    "class_type": "CheckpointLoaderSimple",
    "_meta": {
      "title": "Load Checkpoint"
    }
  },
  "8": {
    "inputs": {
      "seed": 62282230408842,
      "steps": "50",
      "cfg": "7.5",
      "sampler_name": "ttm",
      "scheduler": "ddim_uniform",
      "denoise": 1,
      "noise_mode": "GPU(=A1111)",
      "batch_seed_mode": "incremental",
      "variation_seed": 0,
      "variation_strength": 0,
      "variation_method": "linear",
      "model": ["2", 0],
      "positive": ["9", 0],
      "negative": ["10", 0],
      "latent_image": ["15", 0]
    },
    "class_type": "KSampler //Inspire",
    "_meta": {
      "title": "KSampler (inspire)"
    }
  },
  "9": {
    "inputs": {
      "text": "{positive_prompt}",
      "token_normalization": "length+mean",
      "weight_interpretation": "comfy++",
      "clip": ["2", 1]
    },
    "class_type": "BNK_CLIPTextEncodeAdvanced",
    "_meta": {
      "title": "CLIP Text Encode (Advanced)"
    }
  },
  "10": {
    "inputs": {
      "text": "blurry, low quality, deformed, disfigured",
      "token_normalization": "length+mean",
      "weight_interpretation": "comfy++",
      "clip": ["2", 1]
    },
    "class_type": "BNK_CLIPTextEncodeAdvanced",
    "_meta": {
      "title": "CLIP Text Encode (Advanced)"
    }
  },
  "12": {
    "inputs": {
      "vae_name": "SDXL 1.0/sharpspectrum_vaexl.safetensors"
    },
    "class_type": "VAELoader",
    "_meta": {
      "title": "Load VAE"
    }
  },
  "13": {
    "inputs": {
      "tile_size": 512,
      "samples": ["8", 0],
      "vae": ["12", 0]
    },
    "class_type": "VAEDecodeTiled",
    "_meta": {
      "title": "VAE Decode (Tiled)"
    }
  },
  "15": {
    "inputs": {
      "width": "1024",
      "height": "1024",
      "batch_size": 1
    },
    "class_type": "EmptyLatentImage",
    "_meta": {
      "title": "Empty Latent Image"
    }
  },
  "22": {
    "inputs": {
      "filename_prefix": "{save_prefix}",
      "images": ["13", 0]
    },
    "class_type": "SaveImage",
    "_meta": {
      "title": "Save Image"
    }
  }
}
\end{verbatim}
\paragraph{Encoded Form}
\begin{verbatim}
N2: CheckpointLoaderSimple
[ckpt_name=SDXL 1.0/realvisxlV40_v40LightningBakedvae.safetensors]
N8: KSampler //Inspire
[steps=50,cfg=7.5,scheduler=ddim_uniform,model=N2,--
positive=N9,negative=N10,latent_image=N15]
N9: BNK_CLIPTextEncodeAdvanced
[clip=N2]
N10: BNK_CLIPTextEncodeAdvanced 
[text=blurry, low quality, deformed, disfigured,clip=N2]
N12: VAELoader 
[vae_name=SDXL 1.0/sharpspectrum_vaexl.safetensors]
N13: VAEDecodeTiled 
[samples=N8,vae=N12]
N15: EmptyLatentImage
N22: SaveImage
[images=N13]

\end{verbatim}
\subsection{User study}
To evaluate our method against baselines, we conducted a user study using a structured survey. For the study, we randomly sampled 50 prompts and generated corresponding images with each baseline. From these, we filtered out results which contained unsafe content (e.g., nudity, violence), resulting in 7–11 comparison questions per baseline. These comparisons were aggregated into a survey where participants were shown a prompt and the outputs from \ourmethod{} and one baseline, and asked to select their preferred image.

\begin{figure}[ht]
    \centering
    \includegraphics[width=1.\linewidth]{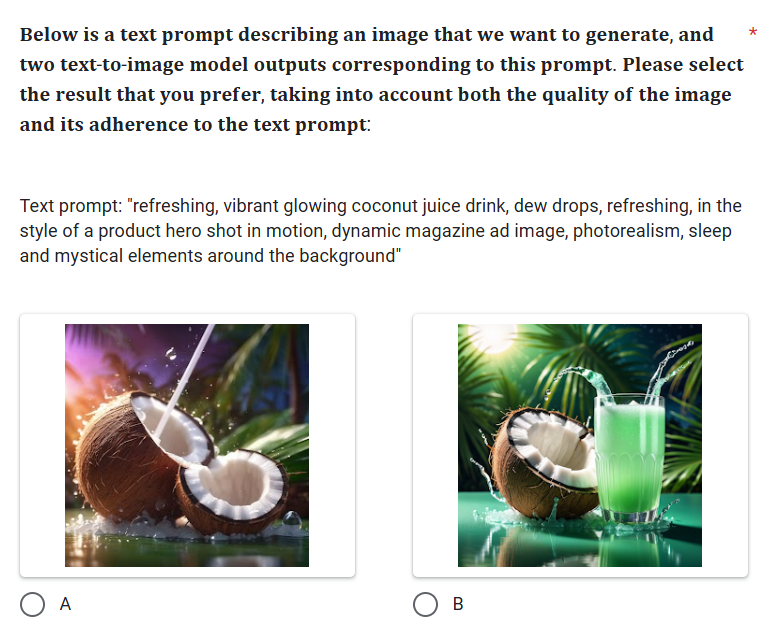}
    \caption{An example question from the user study.}
    \label{fig:user_study_example}
\end{figure}

We collected approximately 200 responses per baseline. 
Figure~\ref{fig:user_study_example}, provides an example of a question from our survey.

\subsection{Reward model generalization}

We conducted an experiment where we re-trained the reward model using half of our original training set, and evaluated the generalization capabilities of our reward model on the remaining half. We further ensured that some graph structures appear only in the hold-out set, so we can evaluate performance on entirely unseen flow graphs.

 The results show good overall performance on the hold-out set, with an $R^2$ score of $0.643$ and a pearson correlation of $0.816$, indicating a strong relationship between predicted and actual values. A more in-depth look shows that: (1) The model generalizes very well to scenarios which contain only parameter or prompt changes compared to what it saw during training (Pearson 0.928). (2) Performance remains good for flows with novel graph structures, but using only seen components (models / blocks) (Pearson 0.667). (3) Performance drops significantly for flows with entirely unseen components, which contain tokens that the reward model has never seen (Pearson 0.152). 

After running this experiment, we also evaluated the GRPO stage using the new reward model. The model trained on the full dataset outperforms the model trained on the partial data (71\% HPSv2 win-rate), showing the benefit of additional data.

\subsection{Implementation details}
\subsubsection{SFT stage}
We implement our model based on a pre-trained Meta Llama3.1- 8B~\cite{grattafiori2024llama}. We used the unsloth~\cite{unsloth2024} library to fine-tune the model using LoRA~\cite{Hu2021LoRALA}. The SFT stage was trained on a single NVIDIA H100 80GB HBM3 GPU for 10 hours.

\paragraph{LoRA Configuration:}
To enable parameter-efficient fine-tuning, we applied LoRA (Low-Rank Adaptation) to the model’s attention and feed-forward layers. The LoRA rank was set to $r = 16$, with an alpha value of $\alpha=16$, and a dropout rate of $0.0$. Target modules included "q\_proj, k\_proj, v\_proj, o\_proj, gate\_proj, up\_proj, down\_proj", as well as "lm\_head" and "embed\_tokens" since we added new tokens to our vocabulary.

\paragraph{Prompt structure}
During the supervised fine-tuning (SFT) stage, the LLM is provided with both the prompt and one of the encoded flows:
\begin{verbatim}
">>> Prompt:
    {p_i}
>>> Flow:
    {f_$}"
\end{verbatim}
In contrast, during the reinforcement learning (RL) fine-tuning stage, only the prompt is given to the LLM, and it is tasked with generating one or more candidate flows. This setup encourages the model to learn to produce the most appropriate flow for each prompt:
\begin{verbatim}
">>> Prompt:
    {p_i}
>>> Flow:"
\end{verbatim}
\subsubsection{Reward model training}
For training the Reward BERT model, we utilized the "answerdotai/ModernBERT-base"~\cite{warner2024smarter} as the foundational architecture. Beyond its improved classification performance, we selected ModernBert because it was trained on sequence lengths that match our expected prompt and encoded-flow format. We used the Adam optimizer with the default parameters and a learning rate of $8e-5$. The maximum sequence length was set to $4096$ tokens, with a batch size of $128$  over $10$ epochs. Fine-tuning was done on a single NVIDIA A100-SXM4-80GB for approximately 4 hours.

\paragraph{Dataset:}
Each data-point consisted of the triplet $(f_i,p_i,s_i)$: flow, prompt and human-preference normalized score. and was inserted to the model in this format:
\begin{verbatim}
"[PROMPT] {p_i} [FLOW] {f_i}".
\end{verbatim}

The model was tasked with prediction the output score $s_i$ for each pair, using an MSE loss.

\subsubsection{GRPO Fine-Tuning Hyperparameters}
Below, we detail the key hyperparameters and configurations used in the GRPO (Group Relative Policy Optimization) fine-tuning stage:

\paragraph{LoRA Configuration:}
To enable parameter-efficient fine-tuning, we applied LoRA (Low-Rank Adaptation) to the model’s attention and feed-forward layers. The LoRA rank was set to $r = 16$, with an alpha value of $\alpha=16$, and a dropout rate of $0.0$. Target modules included "q\_proj, k\_proj, v\_proj, o\_proj, gate\_proj, up\_proj, down\_proj", following best practices for large language model adaptation. Note that this step does not optimize the "lm\_head" or "embed\_tokens" layers as this step aims to further tune the SFT model, which already knows the flow vocabulary. 

\paragraph{Optimization Settings:}
We used the Adam optimizer with a learning rate of $5e-6$, $\beta_1 = 0.9, \beta_2 = 0.99$, and a weight decay of $0.1$. Training was performed with a batch size of $16$ per device.

\paragraph{GRPO-Specific Parameters:}
We used the group size of $4$ (number of generations per prompt for group-based reward calculation). clipping coefficient of $0.2$, max grad norm of $0.5$ and KL-regularization coefficient of $0.2$. We also used generation temperature of $0.9$, and maximal output tokens of $500$.

\paragraph{Training Procedure:}
Fine-tuning was conducted for 2 epochs over the CivitAI promp train set. We trained on a single NVIDIA A100-SXM4-80GB node (8 GPUs) for approximately 10 hours. We used an ensemble of 7 BERT reward models and used their mean as the surrogate reward. we set the uncertainty threshold to 0.08 and set the "uncertain reward value" to $0$. For the prefix-reward mechanism, we used $5$ different Bert models.


